\documentclass{article}
\usepackage[utf8]{inputenc}
\usepackage[T1]{fontenc}
\usepackage{microtype}
\PassOptionsToPackage{hyphens}{url}
\usepackage{hyperref}

\usepackage[preprint]{icml2027}
\usepackage{etoolbox}
\makeatletter
\ifdefined\ispreprint
  \patchcmd{\printAffiliationsAndNotice}
    {\ifdefined\icmlcorrespondingauthor@text
       { }Correspondence to: \icmlcorrespondingauthor@text.
     \else
       {\bf AUTHORERR: Missing \textbackslash{}icmlcorrespondingauthor.}
     \fi}
    {}{}{\PackageError{JevRisk}{Could not omit the preprint correspondence line}{}}
\fi
\makeatother

\usepackage{amsmath,amsfonts,bm}

\def\eqref#1{equation~\ref{#1}}

\def\1{\bm{1}}

\DeclareMathAlphabet{\mathsfit}{\encodingdefault}{\sfdefault}{m}{sl}
\SetMathAlphabet{\mathsfit}{bold}{\encodingdefault}{\sfdefault}{bx}{n}

\usepackage{amsmath}
\usepackage{amssymb}
\usepackage{mathtools}
\usepackage{amsthm}
\usepackage{graphicx}
\usepackage{booktabs}
\usepackage{array}
\usepackage{tabularx}
\usepackage{float}
\usepackage{multirow}
\usepackage{xcolor}
\usepackage{tikz}

\definecolor{exampleback}{HTML}{F1F2F1}
\definecolor{findingback}{HTML}{FFF5D6}
\definecolor{findingframe}{HTML}{D9A92B}
\definecolor{principleback}{HTML}{F2ECFA}
\definecolor{principleframe}{HTML}{8662C4}
\newsavebox{\jevboxcontent}
\newcommand{\jevboxskip}{\medskip}
\newenvironment{jevbox}[3]{\par\jevboxskip\noindent\def\jevboxframe{#1}\def\jevboxback{#2}\setlength{\fboxsep}{6pt}\setlength{\fboxrule}{0.6pt}%
  \begin{lrbox}{\jevboxcontent}\begin{minipage}{\dimexpr\linewidth-2\fboxsep-2\fboxrule\relax}\textbf{#3.}\ \ignorespaces}%
  {\end{minipage}\end{lrbox}\fcolorbox{\jevboxframe}{\jevboxback}{\usebox{\jevboxcontent}}\par\jevboxskip}
\newcounter{example}

\newenvironment{principlebox}[1]{\begin{jevbox}{principleframe}{principleback}{#1}}{\end{jevbox}}
\definecolor{promptback}{HTML}{EEF3FA}
\definecolor{promptframe}{HTML}{7D9BC1}
\newenvironment{promptbox}[1]{\begin{jevbox}{promptframe}{promptback}{#1}\par\vspace{3pt}\noindent\small\ignorespaces}{\end{jevbox}}
\newenvironment{promptspec}[1]{\begin{jevbox}{promptframe}{promptback}{#1}\small\par\smallskip\noindent\begin{tabular}{@{}>{\bfseries\raggedright\arraybackslash}p{2.35cm}@{\hspace{6pt}}>{\raggedright\arraybackslash}p{\dimexpr\linewidth-2.35cm-6pt\relax}@{}}}{\end{tabular}\end{jevbox}}
\usetikzlibrary{arrows.meta,positioning,fit,calc}

\theoremstyle{definition}

\newcommand{\jev}{Jev}

\definecolor{vizblue}{HTML}{2A78D6}
\definecolor{vizaqua}{HTML}{1BAF7A}
\definecolor{vizorange}{HTML}{EB6834}
\definecolor{vizgray}{HTML}{8A8984}

\icmltitlerunning{JevRisk for Financial Risk Assessment}

\begin{document}

\twocolumn[
  \icmltitle{JevRisk: Zero-Shot Financial Risk Assessment with a Language Decision Model}

  \begin{icmlauthorlist}
    \icmlauthor{Yaodong Yang}{swufe}
    \icmlauthor{Li Huang}{swufe}
    \icmlauthor{Junpeng Liang}{abc}
    \icmlauthor{Yi Ma}{sxu}
    \icmlauthor{Hongyao Tang}{tju}
    \icmlauthor{Xingyu Fan}{cuhk}
    \icmlauthor{Jinpeng Li}{cuhk}
    \icmlauthor{Furui Liu}{ucas}
  \end{icmlauthorlist}

  \icmlaffiliation{swufe}{School of Computing and Artificial Intelligence, Southwestern University of Finance and Economics}
  \icmlaffiliation{abc}{Agricultural Bank of China}
  \icmlaffiliation{sxu}{Shanxi University}
  \icmlaffiliation{tju}{Tianjin University}
  \icmlaffiliation{cuhk}{The Chinese University of Hong Kong}
  \icmlaffiliation{ucas}{University of Chinese Academy of Sciences}

  \vskip 0.3in
]
\printAffiliationsAndNotice{}

\begin{abstract}
Banks must monitor every customer's financial risk, such as money laundering and credit default, by interpreting financial activity in the context of the customer's profile. Existing risk monitoring systems of banks are mostly based on expert-designed risk indicators, which apply predefined thresholds to every customer and cannot adapt to their personal situations. Supervised models recognise risk patterns from labelled customers, but each label often needs an expert's review or months of repayment history, which is slow and costly. Financial risk assessment therefore calls for a zero-shot decision system that minimises the need for labels and generalises across customers and risk types. To this end, we present JevRisk, a system for zero-shot financial risk assessment with a language decision model. JevRisk works in two steps. First, following regulatory guidelines in the finance domain, JevRisk computes exact risk indicators with code. Second, the language decision model (e.g., \jev{}) reads these indicators together with the customer's profile and activity, judges the case in context with broad world knowledge, and returns the score of high risk. We evaluate JevRisk on the anti-money laundering task with public synthetic data from Scotiabank and on the credit risk task with real data from Home Credit and Lending Club. On both tasks, JevRisk outperforms rule-based and unsupervised methods, and is on par with or better than supervised methods (e.g., GBDT and TabPFN) trained on about a thousand labelled customers.
\end{abstract}
\section{Introduction}
\label{sec:intro}

Banks must rate the risk of every customer. For example, the rating for anti-money laundering (AML) decides which customers receive enhanced monitoring \citep{fatf2012recommendations,fintrac2021rba}, and the credit assessment decides who receives a loan and at what interest rate and amount \citep{eba2020loan}. Both assessments depend on context, since the same activity can be normal for one customer and a warning sign for another. For example, large cash deposits may fit a restaurant owner but not a student, and a large loan may suit a civil servant with a stable income but not a seasonal worker. Regulators such as the international Financial Action Task Force (FATF), Canada's FINTRAC and the European Banking Authority (EBA) therefore ask banks to judge each customer in context \citep{fatf2014rba,fintrac2024indicators,eba2020loan}.

\begin{figure*}[t]
\centering
\includegraphics[width=\linewidth]{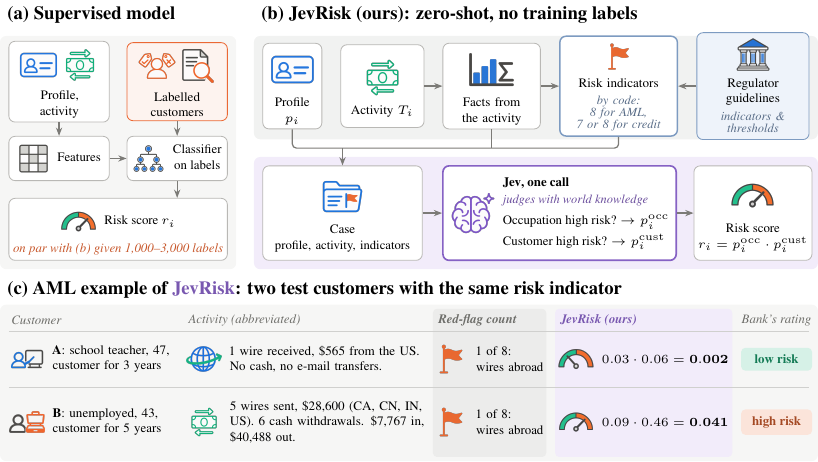}
\caption{\textbf{JevRisk computes risk indicators with code and lets \jev{} rate the customer.} (a) A supervised model learns from labelled customers. (b) JevRisk builds a case from the profile, activity, and indicators. It multiplies the two probabilities returned by \jev{} in one call. (c) Two AML test customers \citep{imi2024data} trigger the same red flag. JevRisk scores the unemployed customer higher.}
\label{fig:framework}
\end{figure*}

Analysts cannot review every customer by hand, so banks rely on automated scoring. Existing risk monitoring systems of banks are mostly based on expert-designed rules that count risk indicators. These rules, however, apply predefined thresholds to every customer and cannot adapt to their personal situations. In Figure~\ref{fig:framework}c, a teacher who received one small wire from abroad and an unemployed customer who sends out five times more money than comes in share the same risk indicator of ``wires abroad''. Such a rule thus rates the two customers alike, yet an analyst would rate them differently. Banks then resort to supervised models, which recognise risk patterns from labelled cases \citep{jullum2020detecting,lessmann2015benchmarking}, but each label often needs an expert's review or months of repayment history, which is slow and costly. Moreover, labels are specific to one risk, so each new risk type needs new labels. Models trained on past labels can also degrade when customers and risk patterns change \citep{dalpozzolo2018credit,lu2019concept}. Financial risk assessment therefore calls for a zero-shot decision system that judges each customer in context, minimises the need for labels, and generalises across risk types.

To this end, we present JevRisk, a system for zero-shot financial risk assessment that uses a language decision model such as \jev{} \citep{typesafe2026jev}. Given a case described in text and a yes/no question, \jev{} returns the probability of ``yes'' in a single call. We argue that such a model fits financial risk assessment for three reasons. First, the model \emph{brings broad knowledge of the world} for generalization, such as which income, spending, and transactions are normal for an occupation, an age, or a country. Second, the model \emph{reads structured text}, which supports different types of tasks. A case file can hold a customer profile, a summary of the activity and risk indicators with definitions and evidence in one input. Third, the model \emph{returns a probability}. The output probabilities rank customers and are directly suitable for financial risk assessment.

However, \jev{} evaluates the risk probability well only when the input states the exact information for the judgment \citep{yang2026code}. When it must read long records and compute sums and counts itself, its accuracy drops and its cost rises (Appendix~\ref{app:ablation}). Therefore, JevRisk works in two steps that separate exact computation from risk judgment in context (Figure~\ref{fig:framework}). First, following regulatory guidelines, JevRisk computes exact risk indicators (FINTRAC's money-laundering red flags for AML and the EBA's early-warning indicators for credit) with code based on the customer's activity and transactions \citep{fintrac2024lctr,fintrac2024eft,eba2020loan,eba2018npe}. Second, the language decision model \jev{} reads these indicators together with the customer's profile and a summary of the activity as structured text, i.e., the case file that an analyst would read, and judges the case in context. In the same call, \jev{} returns the probabilities that the occupation is high risk and that the customer is high risk, and their product is the final risk score for assessment. We test JevRisk on two tasks. For AML, we use the public synthetic data curated by Scotiabank \citep{imi2024data}. For credit risk, we use real data on consumer loans at Home Credit and Lending Club \citep{homecredit2018data,george2019lendingclub}. On both tasks, JevRisk outperforms the label-free baselines, namely counting the same indicators and an isolation forest. Surprisingly, JevRisk is on par with or better than supervised models trained on about a thousand labelled customers, including a gradient-boosted decision tree (GBDT), logistic regression, and the tabular foundation model TabPFN \citep{hollmann2025accurate}. Our contributions are as follows.
\begin{itemize}\setlength{\itemsep}{1pt}\setlength{\topsep}{2pt}
\item \textbf{Application.} To our knowledge, this paper is among the first studies of a language decision model (e.g., \jev{}) applied to financial tasks, i.e., AML customer risk rating and credit risk assessment (Sections~\ref{sec:method} and~\ref{sec:experiments}). We argue that three properties of such a model fit financial risk assessment, namely the broad knowledge, the text input and the probability output.
\item \textbf{Method.} We propose JevRisk, which encodes the risk-based approach of regulators in a zero-shot scoring system. First, code computes the risk indicators that regulators list, based on common practice in finance. Second, the language decision model \jev{} judges the risk from these indicators and the customer's profile, as an analyst would (Section~\ref{sec:method}).
\item \textbf{Evaluation.} We test JevRisk on AML customer risk rating with public synthetic bank data and on credit risk assessment with real data of two lenders. On both tasks, JevRisk outperforms rule-based and unsupervised baselines and is on par with or better than supervised models trained on about a thousand labelled customers (Section~\ref{sec:experiments}).
\end{itemize}
\section{Task and Data}
\label{sec:setting}

\textbf{Task.} For each customer $i$, a bank holds a profile $p_i$ on file and the activity $T_i$ of an observation period, such as transactions or a credit history. A risk-scoring method returns a score $r_i$. The bank then decides which customers to rate high risk and to review or monitor more closely. Rating each customer is thus a one-step decision. The state is $(p_i, T_i)$, and the action is to rate customer $i$ high risk or not. We use \emph{zero-shot} to mean that JevRisk scores customers without training on labelled examples from the target task. It uses unlabelled development records to set thresholds for indicators such as unusually frequent cash deposits.

\textbf{Metric.} We evaluate each method by the AUC. Taking a high-risk customer and a low-risk customer at random, the AUC is the probability that the method scores the high-risk one higher. The AUC fits the task because the bank uses the score as a ranking. Usually, analysts can review only a limited number of customers, so the bank reviews customers in the order of their scores, from the highest to the lowest. Where the bank stops depends on its risk appetite, i.e., how much risk it accepts, and on how many customers its analysts can review. The AUC is therefore suitable, as it measures the whole ranking. We also report precision and recall at the top of the ranking in Appendix~\ref{app:pr}.

\textbf{AML customer risk rating.} We use the public data curated by Scotiabank \citep{imi2024data}. The data are synthetic and include 195,789 customers, each with a gender, an occupation, an age, and the years as a customer, which form the profile $p_i$. The transactions comprise 212,532 cash deposits and withdrawals, 506,451 e-mail money transfers (EMTs, a common Canadian payment between persons) and 67,872 wires with the countries of the sender and receiver. The transactions of a customer form the activity $T_i$. Each customer carries a label, high risk or not, and 2.82\% are high risk. The risk label of the customer is rated by Scotiabank.

\textbf{Credit risk assessment.} We use the public data of Home Credit, a consumer lender \citep{homecredit2018data}. The data are real and anonymised loan applications, 307,511 in all, with the repayment outcome of each. In this task, the customers are the applicants. Each application states the occupation (18 kinds), the income type and the employer's sector. The application also states the education, the family, the housing, the age, the years employed, the income and the requested loan. The credit history holds the credits at other institutions, as reported by a credit bureau, and the previous loans with Home Credit. The previous loans come with the instalments, the card balances and the point-of-sale loans (loans for goods bought in a shop). The application forms the profile $p_i$, and the credit history forms the activity $T_i$. The label comes from Home Credit's repayment records and marks the 8.07\% of applicants who had payment difficulties on the loan. We also use the public data of Lending Club, a US peer-to-peer lender \citep{george2019lendingclub}. We take its 546,018 finished loans of 36 months issued from 2013 to 2015. Each application states a job title in free text, the years employed, the housing, the income and the amount and purpose of the loan. The credit history is a summary of the applicant's credit report, such as the accounts, the balances, the inquiries and the past-due payments. The label marks the loans that were charged off or defaulted, 13.2\% of the loans issued in 2013 and 2014 and 14.9\% of those issued in 2015.

\textbf{Splits.} For the AML data and Home Credit, we split the customers at random into a development set (about 20\%, 38,712 bank customers and 61,694 applicants) and a test set (about 80\%). These data carry no dates of the transactions or the applications, so a split by time is not possible. Lending Club records the issue date of each loan, so we split its loans by time. The 262,992 loans issued in 2013 and 2014 form the development set, and the 283,026 loans issued in 2015 form the test set. JevRisk is thus tested on later loans, as a bank scores new applicants after its past ones. The development set can be regarded as the customer data accumulated by the bank. It supplies the unlabelled customers that the regulators' risk indicators involve (Section~\ref{sec:method}) and the labels of the supervised baselines.
\section{JevRisk for Financial Risk Assessment}
\label{sec:method}

This section describes JevRisk, which works in two steps. First, following regulatory guidelines, code computes exact risk indicators and joins them with the profile and activity into a \emph{case}, the file an analyst would review. Second, \jev{} judges the case in context with one call that asks two questions, and JevRisk combines the two answers into the risk score. Figure~\ref{fig:framework} shows the AML instance.

\begin{principlebox}{Design principle}
JevRisk follows common practice in finance. Code computes the risk indicators that regulators list, because they need exact sums, counts and comparisons with a threshold. The language decision model \jev{} answers the questions that an analyst would ask, because they need knowledge of the world, such as which activity is normal for which occupation.
\end{principlebox}

\subsection{Code computes the risk indicators}
\label{sec:flags}

Regulators publish the indicators that a bank should check, such as money-laundering red flags and the early-warning indicators of credit risk. Code first aggregates the activity of each customer into a few facts and then computes the indicators from these facts. An indicator is present when a fact reaches a threshold. Sometimes guidelines state the threshold. For example, FINTRAC requires a report for a cash deposit of \$10,000 or more, so a single deposit of this size is an indicator. Sometimes guidelines name only an unusual activity, such as frequent cash deposits, without an amount. For such an indicator, we take the 95th percentile over the development customers as a peer-group threshold and flag a customer as having frequent cash deposits if they reach it. The threshold is computed once, without labels or test customers, and then fixed. Banks likewise calibrate monitoring thresholds on the historic activity of the customers \citep{fiau2023monitoring}.

\textbf{AML red flags.} For each customer, code aggregates the transactions by channel (cash, EMTs and wires) into counts, sums, distinct counterparties, the largest wires and the countries of sender and receiver (Appendix~\ref{app:flags}). From these facts, code computes the eight red flags of Table~\ref{tab:flags}, which follow FINTRAC's money-laundering indicators for financial entities \citep{fintrac2024indicators}. For example, a cash deposit of \$10,000 or more triggers a large cash transaction report to FINTRAC \citep{fintrac2024lctr}. A wire to or from another country of \$10,000 or more triggers an electronic funds transfer report \citep{fintrac2024eft}. A transaction that triggers such a report is not suspicious in itself, but draws attention. The peer-group thresholds are set on the development set, i.e., the customer data that a bank has accumulated.

\textbf{Credit early-warning indicators.} For Home Credit, code aggregates the credit history into the credits at other institutions, the previous applications, the instalments, the card balances and the point-of-sale loans (Appendix~\ref{app:credit}). From these facts, code computes the seven indicators of Table~\ref{tab:ewi} in Appendix~\ref{app:credit}, which follow the EBA's early-warning indicators of credit risk \citep{eba2020loan}. For example, the EBA lists early arrears of 5 to 30 days past due as a warning sign, so an applicant with an instalment paid five or more days late gets this indicator \citep{eba2018npe}. The EBA states no amount for many new credits or for a high instalment over income, so these two indicators use the 95th percentile over the development applicants. For Lending Club, code computes eight indicators from the application and the credit report, such as a recent payment past due or many accounts opened in the last year. Where the data allow, they follow the same EBA indicators (Appendix~\ref{app:lending}).

\textbf{The indicator format in the case prompt.} Each indicator enters the case with a definition, a status (present or absent) and evidence, i.e., the number behind the indicator. The entry thus reads much like an alert that an analyst reviews. For example, \texttt{large\_wire\_abroad} of Customer~B in Figure~\ref{fig:framework}c reads ``absent'' with the evidence ``largest wire to or from outside Canada \$4,385''. Absent indicators are kept, so that \jev{} also sees what was checked and did not trigger.

\subsection{JevRisk judges both customer and occupation}
\label{sec:judge}

\jev{} reads the case as structured text, i.e., a JSON state, together with a yes/no question, and returns the probability of ``yes'' \citep{typesafe2026jev}. The case of one customer has three parts (Appendices~\ref{app:interface} and~\ref{app:credit}). The first part holds the profile $p_i$, i.e., the know-your-customer (KYC) data for AML and the application for credit, such as the occupation and the age. The second part summarises the activity $T_i$ in one line per channel, such as ``5 totalling \$28,600, largest \$8,066'' for the wires sent. The third part holds the risk indicators of the corresponding task computed by code.

Under the risk-based approach, a bank rates a customer on several risk factors, such as the type of customer, the geography, and the products used \citep{fatf2012recommendations,fatf2014rba}. Specifically, the occupation is part of the type of customer. Some occupations carry a higher inherent risk, i.e., a higher risk before any activity is considered, such as the owner of a cash-intensive business. For credit, occupation also matters because it shows income stability, e.g., a civil servant versus a seasonal worker. If JevRisk asked only whether the customer is high risk, occupation would be one of many inputs to that answer, and we could not see how much it counts. Therefore, JevRisk asks a separate question on the risk of the occupation, which makes the occupation an explicit risk factor, as in the bank's rating. Both questions share one context and differ only in the last sentence. The AML questions are as follows. The questions for Home Credit and Lending Club are in Appendices~\ref{app:credit} and~\ref{app:lending}, respectively.

\begin{promptbox}{Questions to \jev{} for AML by JevRisk}
\textbf{Context.} You are an expert financial crime investigator at a bank. \texttt{case.customer} is what a retail customer declared when opening the account. \texttt{case.activity} summarises all its cash, e-mail money transfer (EMT) and wire activity over the observed period. Code has checked the activity against money-laundering red flags; the results are in \texttt{case.red\_flags}.\\[3pt]
\textbf{Occupation} ($p^{\mathrm{occ}}_i$). Does the customer's declared occupation, in itself, make the customer a high money-laundering risk?\\[3pt]
\textbf{Customer} ($p^{\mathrm{cust}}_i$). Is this customer a high money-laundering risk?
\end{promptbox}

\jev{} answers both questions in one call. The risk score is the product of the two answers,
\begin{equation}
  r_i = p^{\mathrm{occ}}_i \cdot p^{\mathrm{cust}}_i .
\end{equation}
A customer thus ranks high only if both answers point to risk. Note that the product serves as a ranking score, not as a joint probability. With \jev{}, scoring a customer takes one API call. This brings a bank two advantages. The first is generality. The same JevRisk serves both tasks, and only the indicators, the case fields and the questions change. A new risk type or a new indicator therefore needs no labels and no retraining. The second is cost. Rating all 195,789 AML customers once costs about \$9, rating all 307,511 Home Credit applicants about \$15, and rating all 546,018 Lending Club loans about \$31 (Appendix~\ref{app:cost}). This is much cheaper than calling an advanced general-purpose language model on every customer \citep{typesafe2026jev}.
\section{Experiments}
\label{sec:experiments}

\begin{table*}[t]
\centering
\small
\caption{\textbf{AML customer risk rating of held-out customers without labels.} AUC on 2,000 test customers rated high risk and 2,000 other test customers. Labels is the number of labelled customers that a method uses, with the number rated high risk in brackets, expected for random samples and observed for all development customers. $\Delta\mathrm{AUC}$ is the AUC of the method minus that of JevRisk with \jev{}, with a 95\% bootstrap interval. Table~\ref{tab:main_full} in Appendix~\ref{app:full} gives the full table with 3,000 and 10,000 labels.}
\label{tab:main}
\setlength{\tabcolsep}{4pt}
\fontsize{8.5}{9.5}\selectfont
\begin{tabular*}{\linewidth}{@{\extracolsep{\fill}}lrlcc@{}}
\toprule
Method & Labels (high risk) & Customer data & AUC [95\% interval] & $\Delta\mathrm{AUC}$ \\
\midrule
Red-flag count & none & 4 percentiles & 0.830 [0.818, 0.842] & $-$0.093 [$-$0.102, $-$0.084] \\
Isolation forest & none & all, no labels & 0.898 [0.887, 0.908] & $-$0.025 [$-$0.033, $-$0.017] \\
\midrule
GBDT & 100 (3) & labelled sample & 0.803 [0.795, 0.810] & $-$0.120 [$-$0.127, $-$0.114] \\
GBDT & 300 (8) & labelled sample & 0.888 [0.880, 0.896] & $-$0.035 [$-$0.041, $-$0.029] \\
GBDT & 1,000 (28) & labelled sample & 0.906 [0.899, 0.913] & $-$0.017 [$-$0.023, $-$0.011] \\
GBDT & 38,712 (1,094) & labelled sample & 0.957 [0.951, 0.963] & +0.034 [+0.028, +0.040] \\
\midrule
Logistic regression & 100 (3) & labelled sample & 0.730 [0.724, 0.737] & $-$0.193 [$-$0.200, $-$0.185] \\
Logistic regression & 300 (8) & labelled sample & 0.843 [0.835, 0.851] & $-$0.080 [$-$0.087, $-$0.073] \\
Logistic regression & 1,000 (28) & labelled sample & 0.895 [0.887, 0.903] & $-$0.028 [$-$0.035, $-$0.021] \\
Logistic regression & 38,712 (1,094) & labelled sample & 0.949 [0.942, 0.956] & +0.026 [+0.020, +0.033] \\
\midrule
TabPFN & 100 (3) & labelled sample & 0.848 [0.841, 0.856] & $-$0.075 [$-$0.081, $-$0.068] \\
TabPFN & 300 (8) & labelled sample & 0.905 [0.897, 0.913] & $-$0.018 [$-$0.024, $-$0.011] \\
TabPFN & 1,000 (28) & labelled sample & 0.923 [0.916, 0.931] & +0.000 [$-$0.006, +0.007] \\
TabPFN & 10,000 (282) & labelled sample & 0.934 [0.927, 0.942] & +0.011 [+0.005, +0.018] \\
\midrule
JevRisk (Qwen3.5-9B) & none & 4 percentiles & 0.908 [0.899, 0.917] & $-$0.015 [$-$0.020, $-$0.010] \\
JevRisk (Clef-Flash) & none & 4 percentiles & 0.921 [0.913, 0.929] & $-$0.002 [$-$0.006, +0.003] \\
JevRisk (Jev-Omni) & none & 4 percentiles & 0.873 [0.861, 0.884] & $-$0.050 [$-$0.059, $-$0.042] \\
\textbf{JevRisk (Jev-1.14.0)} & none & 4 percentiles & 0.923 [0.915, 0.931] & -- \\
\bottomrule
\end{tabular*}
\end{table*}

\begin{table*}[t]
\centering
\small
\caption{\textbf{Ablation of the score on the test samples.} AUC on the 4,000 test customers of each data set. $\Delta\mathrm{AUC}$ is the AUC of the score minus that of the product, with a 95\% bootstrap interval. Count + occupation answer averages the percentile ranks of the two scores (Appendix~\ref{app:ablation}). For Home Credit and Lending Club, $p^{\mathrm{occ}}$ takes the neutral value when no occupation is stated, as in the product. Product without indicators removes the indicators and the facts computed for them from the case (Appendix~\ref{app:ablation}).}
\label{tab:ablation}
\setlength{\tabcolsep}{2.5pt}
\fontsize{8.5}{9.5}\selectfont
\begin{tabular*}{\linewidth}{@{\extracolsep{\fill}}lcccccc@{}}
\toprule
 & \multicolumn{2}{c}{AML} & \multicolumn{2}{c}{Home Credit} & \multicolumn{2}{c}{Lending Club} \\
\cmidrule(lr){2-3}\cmidrule(lr){4-5}\cmidrule(l){6-7}
Score & AUC & $\Delta\mathrm{AUC}$ & AUC & $\Delta\mathrm{AUC}$ & AUC & $\Delta\mathrm{AUC}$ \\
\midrule
Indicator count & 0.830 & $-$0.093 [$-$0.102, $-$0.084] & 0.587 & $-$0.041 [$-$0.055, $-$0.026] & 0.574 & $-$0.077 [$-$0.094, $-$0.059] \\
Occupation answer $p^{\mathrm{occ}}$ & 0.840 & $-$0.083 [$-$0.093, $-$0.074] & 0.559 & $-$0.069 [$-$0.084, $-$0.054] & 0.574 & $-$0.077 [$-$0.094, $-$0.060] \\
Customer answer $p^{\mathrm{cust}}$ & 0.897 & $-$0.026 [$-$0.033, $-$0.020] & 0.627 & $-$0.001 [$-$0.012, +0.010] & 0.644 & $-$0.007 [$-$0.015, +0.000] \\
Count + occupation answer & 0.902 & $-$0.021 [$-$0.025, $-$0.016] & 0.604 & $-$0.024 [$-$0.032, $-$0.016] & 0.598 & $-$0.053 [$-$0.068, $-$0.038] \\
Product without indicators & 0.911 & $-$0.012 [$-$0.016, $-$0.008] & 0.611 & $-$0.017 [$-$0.027, $-$0.007] & 0.645 & $-$0.006 [$-$0.014, +0.002] \\
\midrule
\textbf{Product (ours)} & 0.923 & -- & 0.628 & -- & 0.651 & -- \\
\bottomrule
\end{tabular*}
\end{table*}

This section asks two questions. How well does JevRisk rank held-out customers without labels? And how does JevRisk compare with supervised risk models trained on labelled customers?

\subsection{Setup}
\label{sec:setup}

\textbf{Sample.} We score a balanced sample of each test set, with 2,000 high-risk customers and 2,000 other customers drawn at random. For credit, the 2,000 high-risk applicants are those with payment difficulties for Home Credit and those whose loan was charged off or defaulted for Lending Club. The AUC compares a high-risk customer with a low-risk one, so it does not change with the percentage of each class. The balanced sample therefore gives the same expected AUC as the whole test set. All methods score the same 4,000 customers. The AUC is computed on 4,000 sampled customers, so another sample of 4,000 would give a somewhat different AUC. We therefore give a 95\% interval for each AUC by the bootstrap \citep{efron1993introduction}. From the sample, we redraw 2,000 high-risk and 2,000 other customers with replacement and compute the AUC on them. We repeat this 2,000 times, and the interval leaves out the lowest and the highest 2.5\% of these AUCs. To compare a method with JevRisk, we compute the AUC of the method minus the AUC of JevRisk on each of the 2,000 redrawn samples. The interval of the difference leaves out the lowest and the highest 2.5\% of these 2,000 differences. We call the two methods on par if this interval contains zero.

\textbf{Model.} We use \jev{} version \texttt{jev-1.14.0} through the public API, one call per customer. A call reads about 1,120 tokens for AML, 1,180 tokens for Home Credit, and 1,370 tokens for Lending Club. We test three open models on the same cases and questions. For the general-purpose model Qwen3.5-9B \citep{qwen2026qwen35}, we use the first-token probability of “Yes” as in TabLLM \citep{hegselmann2023tabllm} (Appendix~\ref{app:ablation}). The two decision models are Clef-Flash \citep{cloudflare2026clef} (based on Qwen3.5-9B, both questions in one pass) and Jev-Omni \citep{jevomni2026} (based on Gemma-4-12B, one pass per question). JevRisk means JevRisk with \jev{} unless stated otherwise.

\textbf{Baselines.} We compare against three kinds of risk scoring that a bank could use instead.
\begin{itemize}\setlength{\itemsep}{1pt}\setlength{\topsep}{2pt}
\item \emph{Indicator count}, a rule-based score, called red-flag count for AML and early-warning count for credit. The score is the number of indicators that are present, as in scenario-based transaction monitoring, which scores a customer by the rules that the customer triggers \citep{jensen2023fighting,fiau2023monitoring}. The count needs no labels and uses the same indicators as JevRisk.
\item \emph{Isolation forest} \citep{liu2008isolation}, unsupervised anomaly detection. The detector is fitted on all customers without labels and scores how unusual their activity is (Appendix~\ref{app:baselines}).
\item \emph{Supervised risk models}, trained on labelled customers. We use a gradient-boosted decision tree (GBDT) \citep{pedregosa2011scikit,ke2017lightgbm}, logistic regression as the industry standard of credit scoring \citep{lessmann2015benchmarking}, and TabPFN, a recent advanced tabular foundation model pretrained to predict on small data \citep{hollmann2025accurate}.
\end{itemize}

\textbf{Features and labels.} The supervised models read nearly the same facts and profile as JevRisk (Appendix~\ref{app:baselines}). We train each model on $n$ customers drawn at random from the development set, for $n$ = 100, 300, 1,000, 3,000 and 10,000, and average each metric over twenty random training sets, all evaluated on the same 4,000 test customers. The GBDT and logistic regression are also trained on the whole development set (Appendix~\ref{app:baselines}), while TabPFN is able to fit 10,000 samples at most.

\subsection{AML: JevRisk puts the red flags in context}
\label{sec:results}

\textbf{JevRisk outperforms the label-free baselines.} On AML, \textbf{JevRisk achieves an AUC of 0.923}, compared with 0.830 for the red-flag count and 0.898 for isolation forest (Table~\ref{tab:main}). Because analysts can review only a limited number of customers, we compare how many high-risk customers each ranking identifies at the same workload. For example, reviewing the top 5\% of each ranking captures an estimated 49.3\% of all customers rated high risk with JevRisk and 37.7\% with the red-flag count (Table~\ref{tab:pr}). This result suggests that JevRisk could help AML analysts prioritise cases for review. Figure~\ref{fig:framework}c illustrates how customers with the same red-flag count can receive different scores. Customers A and B have the same count, but JevRisk gives the two customers 0.002 and 0.041, in line with the bank's rating. The isolation forest scores how unusual the activity is among all customers, while JevRisk judges whether the activity fits the customer's own KYC profile.

\textbf{Ablation.} On AML, the occupation answer and the indicator count achieve similar AUCs individually (0.840 and 0.830). Averaging their percentile ranks raises AUC to 0.902, suggesting that the two scores capture complementary information. The customer-risk answer alone reaches 0.897, and combining it with the occupation answer raises AUC to 0.923 (Table~\ref{tab:ablation}). Appendix~\ref{app:ablation} compares the product with other ways to combine the two answers. Without the indicators and the facts computed for them, the product drops to 0.911 (Appendix~\ref{app:ablation}). Using Qwen3.5-9B in place of \jev{} gives an AUC of 0.908, compared with 0.923 for \jev{} and 0.830 for the red-flag count (Table~\ref{tab:main}).

\begin{table*}[t]
\centering
\small
\caption{\textbf{Credit risk assessment on held-out applicants.} Each lender's sample has 2,000 bad and 2,000 other applicants. Bad means payment difficulties for Home Credit and a charged-off or defaulted loan for Lending Club. Lending Club's test loans were issued after its development loans. $\Delta\mathrm{AUC}$ is the AUC difference from JevRisk, with a 95\% bootstrap interval. Appendix~\ref{app:full} gives the full results.}
\label{tab:credit}
\setlength{\tabcolsep}{3pt}
\fontsize{8.5}{9.5}\selectfont
\begin{tabular*}{\linewidth}{@{\extracolsep{\fill}}lcccc@{}}
\toprule
 & \multicolumn{2}{c}{Home Credit} & \multicolumn{2}{c}{Lending Club} \\
\cmidrule(lr){2-3}\cmidrule(l){4-5}
Method & AUC [95\% interval] & $\Delta\mathrm{AUC}$ & AUC [95\% interval] & $\Delta\mathrm{AUC}$ \\
\midrule
Early-warning count & 0.587 [0.571, 0.603] & $-$0.041 [$-$0.055, $-$0.026] & 0.574 [0.557, 0.591] & $-$0.077 [$-$0.094, $-$0.059] \\
Isolation forest & 0.556 [0.539, 0.573] & $-$0.072 [$-$0.090, $-$0.054] & 0.544 [0.526, 0.562] & $-$0.107 [$-$0.129, $-$0.085] \\
\specialrule{\lightrulewidth}{0.5pt}{0.5pt}
GBDT (300 labels) & 0.564 [0.557, 0.570] & $-$0.064 [$-$0.081, $-$0.048] & 0.570 [0.563, 0.579] & $-$0.080 [$-$0.095, $-$0.066] \\
GBDT (1,000 labels) & 0.610 [0.601, 0.619] & $-$0.018 [$-$0.034, $-$0.002] & 0.591 [0.582, 0.600] & $-$0.060 [$-$0.074, $-$0.046] \\
GBDT (3,000 labels) & 0.630 [0.620, 0.640] & +0.002 [$-$0.014, +0.018] & 0.609 [0.599, 0.620] & $-$0.042 [$-$0.055, $-$0.028] \\
\specialrule{\lightrulewidth}{0.5pt}{0.5pt}
Logistic regression (300 labels) & 0.587 [0.579, 0.595] & $-$0.041 [$-$0.058, $-$0.024] & 0.594 [0.585, 0.604] & $-$0.056 [$-$0.070, $-$0.042] \\
Logistic regression (1,000 labels) & 0.640 [0.629, 0.651] & +0.012 [$-$0.004, +0.028] & 0.629 [0.616, 0.641] & $-$0.022 [$-$0.035, $-$0.009] \\
Logistic regression (3,000 labels) & 0.673 [0.660, 0.685] & +0.045 [+0.030, +0.060] & 0.653 [0.638, 0.667] & +0.002 [$-$0.011, +0.015] \\
\specialrule{\lightrulewidth}{0.5pt}{0.5pt}
TabPFN (300 labels) & 0.580 [0.573, 0.588] & $-$0.048 [$-$0.064, $-$0.032] & 0.597 [0.586, 0.607] & $-$0.054 [$-$0.068, $-$0.040] \\
TabPFN (1,000 labels) & 0.628 [0.618, 0.637] & +0.000 [$-$0.017, +0.016] & 0.630 [0.617, 0.643] & $-$0.021 [$-$0.033, $-$0.008] \\
TabPFN (3,000 labels) & 0.664 [0.652, 0.677] & +0.036 [+0.020, +0.052] & 0.656 [0.641, 0.672] & +0.006 [$-$0.007, +0.019] \\
\specialrule{\lightrulewidth}{0.5pt}{0.5pt}
JevRisk (Qwen3.5-9B) & 0.617 [0.600, 0.635] & $-$0.011 [$-$0.022, +0.000] & 0.605 [0.588, 0.623] & $-$0.045 [$-$0.060, $-$0.031] \\
JevRisk (Clef-Flash) & 0.614 [0.596, 0.631] & $-$0.014 [$-$0.025, $-$0.002] & 0.624 [0.606, 0.641] & $-$0.027 [$-$0.037, $-$0.017] \\
JevRisk (Jev-Omni) & 0.622 [0.605, 0.639] & $-$0.007 [$-$0.017, +0.004] & 0.630 [0.613, 0.647] & $-$0.021 [$-$0.031, $-$0.011] \\
\textbf{JevRisk (Jev-1.14.0)} & 0.628 [0.611, 0.645] & -- & 0.651 [0.634, 0.667] & -- \\
\bottomrule
\end{tabular*}
\end{table*}

\textbf{JevRisk is on par with supervised models trained on about a thousand labels.} The supervised models learn from the bank's past ratings. With 100 labelled customers, all three models stay far below JevRisk, at 0.730 to 0.848. With 300 labelled customers, all three models are still below, at 0.843 to 0.905. TabPFN is on par at 1,000 labelled customers, at 0.923. Logistic regression and the GBDT are still below at 1,000, at 0.895 and 0.906, and are on par at 3,000. TabPFN is better than JevRisk from 3,000 labelled customers, and all three models are better at 10,000 (Table~\ref{tab:main_full} in Appendix~\ref{app:full}). With all 38,712 development customers, the GBDT and logistic regression reach 0.957 and 0.949. \textbf{JevRisk is thus on par with or better than supervised models trained on about a thousand labelled customers.} Importantly, each of these labels needs an analyst's review. We also report other metrics, such as the average precision and the recall among the top 5\% and 10\% of the ranking, in Table~\ref{tab:pr} in Appendix~\ref{app:pr}.

The AML data are synthetic, and the label is a bank's rating, not an observed outcome. Does JevRisk also work on real customers and on a different risk?

\subsection{Credit: JevRisk transfers to real applicants}
\label{sec:credit}

\textbf{JevRisk transfers to credit risk.} We apply JevRisk to the credit task with the same workflow as in AML. Only the indicators, the case fields, and the questions change. The label is now an observed outcome: payment difficulties on the loan, i.e., a payment late by more than a set number of days on one of the first instalments. On Home Credit, \textbf{JevRisk achieves an AUC of 0.628}, compared with 0.587 for the early-warning indicator count and 0.556 for isolation forest (Table~\ref{tab:credit}). Moreover, JevRisk with Qwen3.5-9B achieves an AUC of 0.617, which is close to JevRisk (\jev{}).

\textbf{JevRisk weighs the indicator against the applicant.} Two test applicants show where the gain over the rule-based indicator count can come from. A 53-year-old labourer, employed for one year, made 5 of 19 instalment payments late and asks for about eleven times the monthly income. A 35-year-old state servant in the police sector, recorded as core staff and employed for 13 years, made 1 of 34 instalment payments late and asks for about eight times the monthly income. Both have one indicator, an instalment paid five or more days late, so the count rates the two alike. \jev{} gives the two applicants 0.52 and 0.20 for difficulty repaying, and 0.36 and 0.05 for the occupation. The final risk scores are 0.187 and 0.010. The labourer is labelled as having payment difficulties, and the other applicant is not (Appendix~\ref{app:credit} shows the case of the labourer). Table~\ref{tab:ablation} shows the ablation for credit.

\textbf{Home Credit.} With 100 or 300 labels, all three supervised models score below JevRisk. At 300 labels, the best reaches 0.587 versus 0.628 for JevRisk (Table~\ref{tab:credit}). Logistic regression and TabPFN are on par at 1,000 labels and better at 3,000. GBDT is on par at 3,000 and better from 10,000 (Table~\ref{tab:credit_full}). For every method, credit risk is harder to rank than AML risk.

\textbf{JevRisk also ranks the later loans of a second lender.} Lending Club lets us test JevRisk on loans issued after the development loans. Its applicants have no repayment history with the lender, so the case holds no arrears with the lender. The customer question thus asks about difficulty repaying the loan, as for Home Credit, but points \jev{} to another early-warning indicator that the EBA lists, a debt increase in other credit institutions \citep{eba2018npe}. On loans issued in 2015, \textbf{JevRisk achieves an AUC of 0.651}, compared with 0.574 for the early-warning indicator count and 0.544 for isolation forest (Table~\ref{tab:credit}). With Qwen3.5-9B, JevRisk reaches 0.605, below \jev{}. At 300 labels, the best supervised AUC is 0.597 versus 0.651 for JevRisk. All three models remain below JevRisk at 1,000 labels. Logistic regression and TabPFN are on par at 3,000. TabPFN surpasses JevRisk at 10,000 labels, and the other two models do so only with all 262,992 development loans (Table~\ref{tab:lending_full}).

\textbf{Replacing \jev{} across tasks.} With either open decision model, JevRisk outperforms the indicator count on all three data sets (Tables~\ref{tab:main} and~\ref{tab:credit}). Clef-Flash approaches \jev{} on AML (0.921 vs.\ 0.923), and Jev-Omni approaches \jev{} on Home Credit (0.622 vs.\ 0.628). Both score below \jev{} on Lending Club.
\section{Related Work}
\label{sec:related}

\textbf{Machine learning for anti-money laundering.} Most banks run rule-based transaction monitoring, whose scenarios raise alerts on amounts, counts and patterns. Analysts review the alerts by hand, and most alerts turn out to be false positives \citep{jensen2023fighting}. Supervised models improve the triage of alerts and transactions when confirmed cases exist \citep{jullum2020detecting}. Graph models learn laundering typologies from labelled transaction networks, as on the Elliptic Bitcoin data \citep{weber2019anti} and on synthetic bank data such as AMLworld \citep{altman2023realistic}. These benchmarks label transactions and hold no KYC profile, so they test transaction monitoring, not customer risk rating. The AML data that we use instead label customers with a risk rating and record each customer's occupation \citep{imi2024data}, so the customer risk factors of the risk-based approach can be tested \citep{fatf2014rba}. Unsupervised detectors such as the isolation forest need no labels \citep{liu2008isolation}, but they score how unusual the activity is, not whether the activity fits the customer's profile.

\textbf{Credit scoring.} Credit scoring is a long-standing task of supervised learning. In a large benchmark, ensemble classifiers rank borrowers better than logistic regression, given labelled outcomes \citep{lessmann2015benchmarking}. All such models need repayment outcomes, which take months to observe. Language models have also been tuned on labelled credit data to score borrowers \citep{feng2023empowering}. Zero-shot language models have been compared with LightGBM on loan default, with feature attributions that diverge from those of the boosted model \citep{almarri2025interpreting}. JevRisk instead lets code compute the facts and the early-warning indicators, and lets \jev{} judge them in context.

\textbf{Pretrained models for tabular and financial data.} TabLLM writes tabular rows as text and lets a language model classify them, which does well with few or no labels \citep{hegselmann2023tabllm}. TabLLM asks one generic question about the row. JevRisk instead gives \jev{} the case that regulators ask for, with the indicators that they list, and asks two questions, about the customer risk and the occupation risk. Tabular foundation models such as TabPFN are pretrained on synthetic tasks and predict from a few labelled rows in one forward pass \citep{hollmann2025accurate}. TabPFN thus still needs labels. Domain models such as BloombergGPT show that financial knowledge can be learned from text \citep{wu2023bloomberggpt}. In AML compliance, language models write first drafts of suspicious activity reports \citep{naik2025coinvestigator}, label the risk of crypto fund flows \citep{lin2025risktagger} and screen transactions zero-shot \citep{chincholikar2026foundation}. These systems generate text or discrete labels. Other work evaluates \jev{} on financial sentiment \citep{deusser2026evaluating} and on Chinese-language tasks, including finance \citep{wang2026chinesejev}. We instead apply a language decision model to financial risk assessment, where its probability ranks customers like a risk score.
\section{Discussion}
\label{sec:discussion}

\textbf{JevRisk fits where labels are few.} On both tasks, JevRisk is on par with or better than supervised models trained on about a thousand labelled customers. The result points to a use where labels are few, such as a new risk type, a new market, a new product or a smaller institution. We have not tested such a transfer, so a bank would need to validate JevRisk on its own customers first. Validation needs labels as well, but these labels check JevRisk instead of training it, and can come from cases that analysts review anyway. Once labels accumulate, supervised models catch up. In AUC, the best supervised model is better than JevRisk from 3,000 labels for AML and Home Credit, and from 10,000 for Lending Club.

\textbf{JevRisk supports model governance.} Two further properties of JevRisk matter for compliance and model governance. First, every indicator follows a regulator's guideline or a field of the credit report, and every amount has a stated source, namely a reporting threshold, an amount in a guideline, a peer percentile or a stated choice of ours. Then the inputs of each rating can be traced and checked by a regulator, an auditor or a model-validation team. The judgment of \jev{} itself, however, is not explained. Second, JevRisk reads the same case as an analyst, so each score comes with the facts that help the bank document the risk assessment. JevRisk only ranks customers for review, and the decision to apply enhanced measures or to decline a loan stays with the analyst.

\textbf{Limitations.} Appendix~\ref{app:limits} discusses the limitations of the data, the method and the evaluation. For example, the AML data are synthetic, so performance on real AML cases remains untested. Both the risk indicators and the questions still require expert design.

\section{Conclusion}
\label{sec:conclusion}

In this paper, we presented JevRisk, a system for zero-shot financial risk assessment with a language decision model. It works in two steps by encoding the domain-based risk indicators of financial regulators into the scoring workflow. First, code computes the risk indicators listed in regulatory guidelines, as in common practice in banks. Second, the language decision model \jev{} judges the risk from these indicators and the customer's profile, as an analyst would, and returns the score of high risk. Without any labels, JevRisk achieves an AUC of 0.923 on AML customer risk rating, and 0.628 and 0.651 on credit risk assessment with two lenders. On both tasks, JevRisk outperforms rule-based and unsupervised baselines and performs on par with or better than supervised models trained on about a thousand labelled customers. These results suggest that language decision models are a promising tool for financial risk assessment, as their broad knowledge, text input and probability output fit the task. Future work includes evaluating JevRisk on real bank data with confirmed laundering cases and combining its score with supervised models as labels accumulate.

\section*{Impact Statement}
All three data sets are public. The AML data are synthetic, and the two credit data sets are anonymised. The case shown to \jev{} leaves out the gender. We do not test JevRisk for bias across protected groups (Appendix~\ref{app:limits}). JevRisk ranks customers for review, and the decision to apply enhanced measures or to decline a loan stays with an analyst.

\section*{Reproducibility Statement}
Section~\ref{sec:judge} and Appendices~\ref{app:interface}, \ref{app:credit} and~\ref{app:lending} give the exact \jev{} questions and the fields of the case. Appendices~\ref{app:flags}, \ref{app:credit} and~\ref{app:lending} give the computation of every indicator. The AML data are public in a copy that a competition team shares under the MIT license \citep{imi2024data}. The Home Credit data are the public data of a Kaggle competition \citep{homecredit2018data}. We use a public copy of these data, which we checked against the published row counts. The Lending Club data are a public Kaggle data set under the CC0 license \citep{george2019lendingclub}. The \jev{} answers of all reported experiments are stored in raw run files, and every score of logistic regression and TabPFN and every answer of Qwen3.5-9B are stored as well. TabPFN uses version 2.0.9 of the \texttt{tabpfn} package with the public v2 classifier weights. Scripts recompute every number of Tables~\ref{tab:main}, \ref{tab:ablation}, \ref{tab:credit}, \ref{tab:stability}, \ref{tab:raw}, \ref{tab:llm}, \ref{tab:main_full}, \ref{tab:credit_full}, \ref{tab:lending_full} and~\ref{tab:pr} from these files and the data. We will release the code, the raw runs, the stored scores and these scripts. \jev{} is accessed through a public API, so its outputs can change across versions.

\section*{AI Use Statement}
The authors conceived JevRisk, designed the experiments, selected the data sets, risk indicators and baselines, and decided what to report. Claude Code (Anthropic) wrote and ran experimental code under the authors' direction and drafted parts of the manuscript. OpenAI Codex assisted with manuscript review. An OpenAI image generation tool created the illustrative icons in Figure~\ref{fig:framework}. The authors revised the manuscript.

\jev{} produced the scores in the main experiments. Qwen3.5-9B, Clef-Flash and Jev-Omni were evaluated as alternative scoring models. Section~\ref{sec:experiments} and Appendices~\ref{app:interface} and~\ref{app:ablation} describe how these models were used. We evaluate the scores against held-out labels, and the scripts described in the Reproducibility Statement recompute the metrics from stored outputs.

\clearpage
\appendix
\onecolumn
\section{The Jev Interface and the Case File}
\label{app:interface}

Every call is one request to the TypeSafe System One API with the model \texttt{jev-1.14.0} \citep{typesafe2026jev}. A request carries a JSON \texttt{state}, \texttt{\{"case": ...\}} with the case file of one customer, and the two yes/no questions of Section~\ref{sec:judge}. The response gives the probability of ``yes'' for each question. The AML context says that all three KYC attributes were declared at account opening, although the bank, not the customer, records the years as a customer. We keep the prompt as run, since this wording does not change the facts that \jev{} reads.

\begin{promptspec}{Case file of one customer}
\texttt{customer} & \texttt{occupation}, \texttt{age}, \texttt{years\_as\_customer}, the KYC attributes in the data. Name and gender are not shown. \\[2pt]
\texttt{activity} & One line per channel: \texttt{cash\_deposits}, \texttt{cash\_withdrawals} (number and total), \texttt{emt\_sent}, \texttt{emt\_received} (number, total and distinct counterparties), \texttt{wires\_sent}, \texttt{wires\_received} (number, total and largest), and \texttt{wire\_countries}. \\[2pt]
\texttt{red\_flags} & The eight flags of Table~\ref{tab:flags}, each with \texttt{definition}, \texttt{status} (present or absent) and \texttt{evidence}.
\end{promptspec}

An excerpt of the case file of Customer~B in Figure~\ref{fig:framework}c reads as follows, with two of the eight red flags. The other six have the same form.

{\footnotesize
\begin{verbatim}
{"customer": {"occupation": "Unemployed", "age": 43, "years_as_customer": 5},
 "activity": {"cash_deposits": "0 totalling $0",
   "cash_withdrawals": "6 totalling $8,965",
   "emt_sent": "3 totalling $2,923 to 3 receivers",
   "emt_received": "5 totalling $1,818 from 5 senders",
   "wires_sent": "5 totalling $28,600, largest $8,066",
   "wires_received": "2 totalling $5,949, largest $4,450",
   "wire_countries": "CA, CN, IN, US"},
 "red_flags": {
   "international_wires": {"definition": "wires to or from outside Canada",
     "status": "present", "evidence": "3 wires to or from outside Canada"},
   "pass_through": {"definition": "money in and out within 10% of each
       other, on more than the median volume",
     "status": "absent", "evidence": "$7,767 in and $40,488 out"},
   ...}}
\end{verbatim}
}

\section{Red-Flag Computation}
\label{app:flags}

\begin{table}[htbp]
\centering
\small
\caption{\textbf{AML red flags computed by code.} A threshold comes from a FINTRAC report, i.e., the large cash transaction report (LCTR) or the electronic funds transfer report (EFTR) \citep{fintrac2024lctr,fintrac2024eft}, or from a percentile over the development customers. Flagged is the percentage of all customers with the flag.}
\label{tab:flags}
\setlength{\tabcolsep}{3pt}
\footnotesize
\begin{tabularx}{\linewidth}{@{}l>{\raggedright\arraybackslash}Xlr@{}}
\toprule
Red flag & Present if & Threshold & Flagged \\
\midrule
\texttt{structuring} & two or more cash deposits between \$9,000 and \$10,000 & LCTR & 0.5\% \\
\texttt{large\_cash} & a single cash deposit of \$10,000 or more & LCTR & 3.0\% \\
\texttt{large\_wire\_abroad} & a single wire to or from abroad of \$10,000 or more & EFTR & 3.7\% \\
\texttt{international\_wires} & any wire to or from abroad & -- & 9.5\% \\
\texttt{frequent\_cash} & cash deposits at or above the 95th percentile ($\geq$11) & percentile & 0.9\% \\
\texttt{emt\_funnel} & distinct EMT senders at or above the 95th percentile ($\geq$9) & percentile & 4.0\% \\
\texttt{emt\_fan\_out} & distinct EMT receivers at or above the 95th percentile ($\geq$7) & percentile & 4.5\% \\
\texttt{pass\_through} & money in and out within 10\%, above the median volume & percentile & 2.5\% \\
\bottomrule
\end{tabularx}
\end{table}

Code aggregates the transactions of each customer by channel, as a transaction-monitoring system would. For cash, code counts deposits and withdrawals, sums the amounts and keeps the largest deposit. For EMTs, code counts the transfers sent and received, sums the amounts and counts the distinct counterparties. For wires, code counts and sums the wires sent and received, and keeps the largest of each and the largest wire to or from abroad. For geographic risk, code keeps the countries of sender and receiver. The countries in the data are Australia, Canada, China, Georgia, India, Mexico, Russia, Saudi Arabia, the United Kingdom and the United States. Money in sums the cash deposits, EMTs received and wires received, and money out sums the cash withdrawals, EMTs sent and wires sent. The activity is sparse. Half of the active customers have three transactions or fewer.

The flags of Table~\ref{tab:flags} then follow. Several deposits just below the reporting amount of \$10,000 suggest structuring, i.e., splitting the cash to avoid the report \citep{fintrac2024indicators}. \texttt{structuring} counts cash deposits of at least \$9,000 and below \$10,000. The peer-group thresholds are 95th percentiles over the development customers with at least one transaction in the channel, namely 11 cash deposits, 9 distinct EMT senders and 7 distinct EMT receivers. \texttt{pass\_through} requires that money in and money out differ by at most 10\% of money in, i.e., that the money passes through the account. The sum of the two must also exceed the median over active development customers (\$1,242), so that small accounts do not trigger the flag. We fix these four values before the test. Taken over all customers instead, the percentiles would be the same. The median would be \$1,237, which would change the flags of only 8 of the 195,789 customers and of no sampled customer.

\textbf{Choices and limits.} FINTRAC states no amount for four indicators and describes them qualitatively, for example unrelated parties who send e-mail money transfers to the same beneficiary. These four flags therefore take peer-group thresholds. They are also proxies. For example, the data show how many distinct senders a customer has, but not whether the senders are related. The remaining parameters, such as the \$9,000 lower bound of \texttt{structuring}, the 95th percentile and the 10\% tolerance of \texttt{pass\_through}, are our choices. Finally, the \texttt{international\_wires} flag records a fact, not a risk. None of the ten countries in the data is on a FATF list of high-risk jurisdictions, so we leave the geographic risk of the countries to \jev{}.

The peer-group thresholds need few customers. Table~\ref{tab:stability} takes the thresholds over random subsets of the development customers. With 1,000 customers the thresholds vary, and a typical draw changes the flags of 274 of the 4,000 test customers. Yet the AUC of JevRisk stays between 0.921 and 0.925 in all draws, while the AUC of the red-flag count ranges from 0.802 to 0.845.

\begin{table}[h]
\centering
\small
\caption{\textbf{Peer-group thresholds from fewer customers.} For each number of customers, we take 20 random draws of development customers. Thresholds and AUCs give the range over the draws. Test customers changed is the number of the 4,000 test customers whose flags differ from those with all development customers, as the median over the draws with the maximum in brackets. Only these cases were sent to \jev{} again, once each.}
\label{tab:stability}
\setlength{\tabcolsep}{4pt}
\begin{tabular*}{\linewidth}{@{\extracolsep{\fill}}rccccrcc@{}}
\toprule
Customers & Cash & EMT & EMT & Median & Test customers & AUC & AUC \\
& deposits & senders & receivers & volume & changed & JevRisk & red-flag count \\
\midrule
1,000 & 9--12 & 8--10 & 7--8 & \$1,050--\$1,414 & 274 (526) & 0.921--0.925 & 0.802--0.845 \\
3,000 & 10--12 & 8--10 & 7--8 & \$1,160--\$1,401 & 72 (335) & 0.922--0.924 & 0.815--0.842 \\
10,000 & 10--11 & 9 & 7--8 & \$1,176--\$1,300 & 0 (273) & 0.922--0.923 & 0.817--0.832 \\
all 38,712 & 11 & 9 & 7 & \$1,242 & 0 (0) & 0.923 & 0.830 \\
\bottomrule
\end{tabular*}
\end{table}

\section{Credit Risk Assessment}
\label{app:credit}

\textbf{Indicators.} Code aggregates the records of each applicant. For the credits at other institutions, code counts the credits and the active credits, and sums the debt of the active credits. A credit counts as past due if the bureau records an overdue amount, overdue days or a late month in the monthly balance. Code also counts the credits recorded as bad debt or sold and the credits opened in the last year. For the previous loans with Home Credit, code counts the applications, approved and refused. Code also counts the payment records of instalments and the records paid five or more days late, and keeps the largest delay. An instalment can be paid in several payments, so the records outnumber the instalments. The case and Section~\ref{sec:credit} count these records as instalment payments. For the card, code counts the months and the months over the limit, and keeps the largest days past due of the card and point-of-sale loans.

Table~\ref{tab:ewi} gives the seven indicators. They follow the borrower-level monitoring metrics in Annex~3 of the EBA guidelines on non-performing exposures, i.e., loans more than 90 days past due or unlikely to be repaid in full \citep{eba2018npe}. The annex gives 5 to 30 days past due as an example of early arrears. We take five or more days for early arrears and more than 30 days for serious arrears. The indicators are proxies, since the data do not record every item of the annex. For example, a credit sold by the lender stands for a non-performing classification, and a count of new credits for an increase in debt. A high instalment over income stands for an increase of this ratio, which one snapshot of the data cannot show. The two peer-group thresholds, 95th percentiles over the development applicants, are 3 credits opened in the last year and an instalment of 35.5\% of the income.

\begin{table}[h]
\centering
\small
\caption{\textbf{Credit early-warning indicators computed by code.} Each indicator is a proxy for a borrower-level monitoring metric in Annex~3 of the EBA guidelines on non-performing and forborne exposures \citep{eba2018npe}. Where the annex states no amount, the threshold is the 95th percentile over the development applicants. Flagged is the percentage of all applicants with the indicator. Many applicants share the count at a threshold, so such ties can raise this percentage above 5\%.}
\label{tab:ewi}
\setlength{\tabcolsep}{4pt}
\begin{tabularx}{\linewidth}{@{}l>{\raggedright\arraybackslash}X>{\raggedright\arraybackslash}Xr@{}}
\toprule
Indicator & Present if & EBA monitoring metric & Flagged \\
\midrule
\texttt{past\_due\_elsewhere} & a credit at another institution past due & past due in other credit institutions & 11.0\% \\
\texttt{bad\_debt\_elsewhere} & a credit at another institution recorded as bad debt or sold & other non-performing classifications in other credit institutions & 1.7\% \\
\texttt{new\_credit\_elsewhere} & credits opened at other institutions in the last year at or above the 95th percentile (3 or more) & debt increase in other credit institutions & 9.7\% \\
\texttt{early\_arrears} & an instalment with Home Credit paid 5 or more days late & early arrears (e.g.\ 5--30 days past due) & 34.1\% \\
\texttt{serious\_arrears} & a previous loan with Home Credit more than 30 days past due & consumer credit days past due & 6.7\% \\
\texttt{overdraft\_exceeded} & a month with the card balance above the credit limit & months with any overdraft exceeded & 12.2\% \\
\texttt{instalment\_to\_income} & monthly instalment over monthly income at or above the 95th percentile (35.5\% or more) & increase in loan instalment over the payroll ratio & 5.0\% \\
\bottomrule
\end{tabularx}
\end{table}

\textbf{Questions.} The credit questions share the form of the AML questions. The role becomes a credit risk analyst at a consumer lender, and the risk becomes difficulty repaying the loan. The box gives the exact text sent to \jev{}.

\begin{promptbox}{Questions to \jev{} for credit by JevRisk}
\textbf{Context.} You are an expert credit risk analyst at a consumer lender. \texttt{case.applicant} is what an applicant stated in a loan application. \texttt{case.history} summarises the applicant's credits at other institutions (credit bureau) and previous loans with this lender. Code has checked the history against early-warning indicators of the European Banking Authority; the results are in \texttt{case.red\_flags}.\\[3pt]
\textbf{Occupation} ($p^{\mathrm{occ}}_i$). Do the applicant's occupation and income type, in themselves, make difficulty repaying the loan likely?\\[3pt]
\textbf{Customer} ($p^{\mathrm{cust}}_i$). Will this applicant have difficulty repaying the loan?
\end{promptbox}

\textbf{Data.} Payment difficulties mean a payment late by more than a set number of days on one of the first instalments. The data also hold three external credit scores. We leave these scores out of every method, so that no method reads a ready-made score. The occupation is missing for 31\% of the applicants, while every AML customer states an occupation. For these applicants, the occupation question has no occupation to judge. \jev{} still rates them as risky (median $p^{\mathrm{occ}}$ 0.21 on development applicants, against 0.14 for applicants with a stated occupation), so the answer reflects the missing field, not an occupation. For these applicants, JevRisk therefore sets $p^{\mathrm{occ}}_i$ to a neutral value, i.e., the median of $p^{\mathrm{occ}}$ over the applicants who state an occupation in a sample of 800 development applicants, which is 0.14. Such an applicant thus counts as having a typical occupation, and the customer answer alone ranks the applicant.

\textbf{Case.} The case has the same three parts as for AML. \texttt{applicant} holds the fields of the application, with ``not stated'' for a missing occupation and without the gender. Within \texttt{applicant}, \texttt{loan} gives the credit amount, the monthly instalment and the price of the goods that the loan pays for. \texttt{history} holds one line each for the credits at other institutions, the previous applications, the instalments, the credit card and the cash and point-of-sale loans, and \texttt{red\_flags} holds the seven indicators of Table~\ref{tab:ewi}. An excerpt of the case of the labourer in Section~\ref{sec:credit} reads as follows, with two of the seven indicators. The other five have the same form.

{\footnotesize
\begin{verbatim}
{"applicant": {"occupation": "Laborers", "income_type": "Working",
   "organization_type": "Self-employed",
   "education": "Secondary / secondary special",
   "family_status": "Civil marriage", "children": 1, "age": 53,
   "years_employed": "1 years", "housing": "House / apartment",
   "owns_car": true, "owns_realty": false,
   "loan": "Cash loans, credit 1,288,350, monthly instalment 37,800,
            goods price 1,125,000",
   "monthly_income": "112,500"},
 "history": {"other_institutions": "0 credits, 0 active with debt 0",
   "previous_applications": "2 with this lender, 2 approved, 0 refused",
   "instalments": "19 paid, 5 of them 5 or more days late, latest by 25 days",
   "credit_card": "0 months, 0 over the limit",
   "cash_and_pos_loans": "16 months, at most 0 days past due"},
 "red_flags": {
   "early_arrears": {"definition": "an instalment of a previous loan with
       this lender paid 5 or more days late",
     "status": "present", "evidence": "5 of 19 instalments paid 5 or more
       days late"},
   "instalment_to_income": {"definition": "monthly instalment of the
       requested loan over monthly income among the top 5% of applicants",
     "status": "absent", "evidence": "monthly instalment 34% of monthly
       income"},
   ...}}
\end{verbatim}
}

\section{Lending Club}
\label{app:lending}

\textbf{Data.} We use the accepted loans of the public copy \citep{george2019lendingclub} and keep the loans of 36 months that were fully paid, charged off or defaulted. A loan is bad if it was charged off or defaulted. We leave the Lending Club grade, the interest rate, the instalment and the FICO score out of every method, so that no method reads a ready-made score. The instalment follows from the amount, the term and the interest rate, so it would reveal the rate. The loan description is empty for loans from 2014 on and is not used.

\textbf{Indicators.} Table~\ref{tab:lc_flags} gives the eight indicators. The applicants have no repayment history with Lending Club itself, so the two arrears indicators of Home Credit have no counterpart. Four indicators use the 95th percentile over the development loans, namely 5 or more accounts opened in the last 12 months, 3 or more inquiries in the last 6 months, a debt-to-income ratio of 31.0\% or more and a loan of 39.6\% or more of the annual income.

\begin{table}[h]
\centering
\small
\caption{\textbf{Lending Club indicators computed by code.} The EBA column gives the borrower-level monitoring metric of Annex~3 \citep{eba2018npe} that an indicator stands for, and -- marks an indicator without a counterpart in the annex. Flagged is the percentage of all 546,018 loans with the indicator.}
\label{tab:lc_flags}
\setlength{\tabcolsep}{4pt}
\begin{tabularx}{\linewidth}{@{}l>{\raggedright\arraybackslash}X>{\raggedright\arraybackslash}Xr@{}}
\toprule
Indicator & Present if & EBA monitoring metric & Flagged \\
\midrule
\texttt{past\_due\_recently} & a payment 30 or more days past due in the last 2 years & past due in other credit institutions & 20.0\% \\
\texttt{bad\_debt\_elsewhere} & an amount in collections, or a credit charged off in the last 12 months & other non-performing classifications in other credit institutions & 15.9\% \\
\texttt{public\_record} & a derogatory public record, such as a judgment or a bankruptcy & -- & 17.6\% \\
\texttt{new\_credit\_recently} & accounts opened in the last 12 months at or above the 95th percentile (5 or more) & debt increase in other credit institutions & 7.8\% \\
\texttt{many\_inquiries} & credit inquiries in the last 6 months at or above the 95th percentile (3 or more) & -- & 5.5\% \\
\texttt{high\_utilisation} & revolving credit used at 90\% or more of the limit & months with any overdraft exceeded & 6.3\% \\
\texttt{high\_debt\_to\_income} & monthly debt payments over monthly income at or above the 95th percentile (31.0\% or more) & increase in loan instalment over the payroll ratio & 7.1\% \\
\texttt{high\_loan\_to\_income} & loan amount over annual income at or above the 95th percentile (39.6\% or more) & -- & 5.4\% \\
\bottomrule
\end{tabularx}
\end{table}

\textbf{Questions.} The occupation question has the same form as for Home Credit. The applicants have no repayment history with Lending Club, so the customer question asks about an early-warning indicator that the EBA lists, a debt increase in other credit institutions, here the accounts opened and the credit inquiries of the last months. The box gives the exact text sent to \jev{}.

\begin{promptbox}{Questions to \jev{} for Lending Club by JevRisk}
\textbf{Context.} You are an expert credit risk analyst at a consumer lender. \texttt{case.applicant} is what an applicant stated in a loan application. \texttt{case.credit\_history} summarises the applicant's credit report. Code has checked the application and the credit report against early-warning indicators of credit risk; the results are in \texttt{case.red\_flags}.\\[3pt]
\textbf{Occupation} ($p^{\mathrm{occ}}_i$). Does the applicant's occupation, in itself, make difficulty repaying the loan likely?\\[3pt]
\textbf{Customer} ($p^{\mathrm{cust}}_i$). Does the applicant's recent credit seeking, i.e. the credit accounts opened and the credit inquiries in the last months, make difficulty repaying the loan likely?
\end{promptbox}

\textbf{Case.} The case has the same three parts as for Home Credit. \texttt{applicant} holds the fields of the application, with ``not stated'' for a missing job title. \texttt{credit\_history} holds twelve fields of the credit report, and \texttt{red\_flags} holds the eight indicators of Table~\ref{tab:lc_flags}. Code adds to each number of \texttt{applicant} and \texttt{credit\_history} its percentile among the development loans, so that \jev{} sees how the number compares with other applicants. Of all loans, 6.7\% state no job title. For these loans, JevRisk sets $p^{\mathrm{occ}}_i$ to a neutral value, i.e., the median of $p^{\mathrm{occ}}$ over the 734 loans with a job title in a sample of 800 development loans, which is 0.06. An excerpt of the case of a test applicant reads as follows, with six of the twelve credit-history fields and two of the eight indicators.

{\footnotesize
\begin{verbatim}
{"applicant": {"occupation": "Pastry Cook", "years_employed": "< 1 year",
   "home_ownership": "rent",
   "annual_income": "$33,113 (higher than 12% of applicants)",
   "income_verified": "verified",
   "loan": "$10,000 over 36 months for small business (higher than
            46% of applicants)"},
 "credit_history": {
   "years_since_first_credit_line": "18 (higher than 68% of applicants)",
   "open_accounts": "14 (higher than 76% of applicants)",
   ...
   "accounts_opened_last_24_months": "8 (higher than 91% of applicants)",
   ...
   "debt_to_income": "19.4% (higher than 61% of applicants)",
   "inquiries_last_6_months": "2 (higher than 87% of applicants)",
   "accounts_opened_last_12_months": "5 (higher than 96% of applicants)"},
 "red_flags": {
   "new_credit_recently": {"definition": "credit accounts opened in the
       last 12 months among the top 5% of applicants",
     "status": "present", "evidence": "5 accounts opened in the last 12
       months"},
   "many_inquiries": {"definition": "credit inquiries in the last 6
       months among the top 5% of applicants",
     "status": "absent", "evidence": "2 inquiries in the last 6 months"},
   ...}}
\end{verbatim}
}

\section{Ablations}
\label{app:ablation}

Table~\ref{tab:ablation} in Section~\ref{sec:results} splits the score into the parts of the single \jev{} call. All rows except the product without indicators use the answers of the main result on the same test sample, so they take no further calls. Count + occupation answer averages the percentile ranks of the indicator count and of $p^{\mathrm{occ}}$, which puts the two scores on one scale. The rule adds an occupation risk to the count, as a weighted scorecard of risk factors would.

\textbf{Without indicators.} To test what the indicators add, we remove them from the case, together with the facts that code computes only for them, such as the cash deposits between \$9,000 and \$10,000 or the credits opened in the last year. The case then holds the profile and the activity summary, and the context no longer mentions the indicators. The two questions, the product score and the test sample stay the same, and the neutral value of $p^{\mathrm{occ}}$ is taken from this run. Without the indicators, the AUC is lower for AML and Home Credit, by 0.012 and 0.017, and lower but on par for Lending Club, by 0.006 (Table~\ref{tab:ablation}). The facts computed for the indicators thus add accuracy. For Lending Club, the summary still holds the accounts opened and the inquiries, which the customer question asks about.

\textbf{Other combinations of the two answers.} Table~\ref{tab:combine} replaces the product with other ways to combine the same two answers. The product ranks customers as the geometric mean of the two answers does, i.e., with equal weights. The mean, the noisy-OR and a weight of 0.75 on the customer answer stay within 0.007 of the product on all three data sets. Only for Home Credit are the mean and the noisy-OR better than the product, by 0.006 and 0.007. A weight of 0.75 on the occupation answer is below the product on all three data sets. For AML, the gain over the customer answer alone thus comes from asking both questions, not from the product itself.

\begin{table}[h]
\centering
\small
\caption{\textbf{Other combinations of the two answers on the test samples.} AUC on the 4,000 test customers of each data set. $\Delta\mathrm{AUC}$ is the AUC of the score minus that of the product, with a 95\% bootstrap interval. For Home Credit and Lending Club, $p^{\mathrm{occ}}$ takes the neutral value when no occupation is stated, as in the product.}
\label{tab:combine}
\setlength{\tabcolsep}{2.5pt}
\fontsize{8.5}{9.5}\selectfont
\resizebox{\linewidth}{!}{\begin{tabular}{@{}lcccccc@{}}
\toprule
 & \multicolumn{2}{c}{AML} & \multicolumn{2}{c}{Home Credit} & \multicolumn{2}{c}{Lending Club} \\
\cmidrule(lr){2-3}\cmidrule(lr){4-5}\cmidrule(l){6-7}
Score & AUC & $\Delta\mathrm{AUC}$ & AUC & $\Delta\mathrm{AUC}$ & AUC & $\Delta\mathrm{AUC}$ \\
\midrule
Mean of the two answers & 0.921 & $-$0.002 [$-$0.004, +0.000] & 0.635 & +0.006 [+0.001, +0.013] & 0.649 & $-$0.002 [$-$0.005, +0.001] \\
Noisy-OR, $1-(1-p^{\mathrm{occ}})(1-p^{\mathrm{cust}})$ & 0.919 & $-$0.004 [$-$0.007, $-$0.001] & 0.635 & +0.007 [+0.000, +0.014] & 0.648 & $-$0.002 [$-$0.006, +0.001] \\
Weighted, $(p^{\mathrm{occ}})^{0.25}(p^{\mathrm{cust}})^{0.75}$ & 0.916 & $-$0.007 [$-$0.010, $-$0.004] & 0.634 & +0.006 [$-$0.000, +0.013] & 0.650 & $-$0.001 [$-$0.006, +0.003] \\
Weighted, $(p^{\mathrm{occ}})^{0.75}(p^{\mathrm{cust}})^{0.25}$ & 0.903 & $-$0.020 [$-$0.025, $-$0.016] & 0.603 & $-$0.025 [$-$0.034, $-$0.017] & 0.633 & $-$0.018 [$-$0.026, $-$0.011] \\
Customer answer $p^{\mathrm{cust}}$ & 0.897 & $-$0.026 [$-$0.033, $-$0.020] & 0.627 & $-$0.001 [$-$0.012, +0.010] & 0.644 & $-$0.007 [$-$0.015, +0.000] \\
\midrule
\textbf{Product (ours)} & 0.923 & -- & 0.628 & -- & 0.651 & -- \\
\bottomrule
\end{tabular}\unskip}
\end{table}

\textbf{Reading raw records.} To test what the code step adds, we let \jev{} read the raw credit records instead of the facts and the indicators. We use Home Credit, since its records are long, with a median of 56 and up to 553 per applicant. The case holds the same application, then one line per raw record, without the summary and the indicators. The records are each credit at another institution with its monthly status, each previous application, each instalment payment, each credit card month and each cash or point-of-sale loan month. The two questions, the product score and the test sample stay the same. The neutral value of $p^{\mathrm{occ}}$ is again the median over the development applicants who state an occupation, now taken from this run. Table~\ref{tab:raw} shows the result. Overall, the AUC is 0.024 lower than that of JevRisk. The gap is 0.005 for the applicants with at most 30 records, 0.029 for those with 31 to 100 and 0.023 for those with more than 100. The raw records also take about three times the tokens of the case with facts and indicators, and the longest take 15 times. The code step therefore adds accuracy, and it keeps the input of each call short and the same across applicants.

\begin{table}[h]
\centering
\small
\caption{\textbf{\jev{} on raw credit records against JevRisk on the test sample.} AUC on the 4,000 test applicants, and the AUC of raw records minus that of JevRisk with a 95\% bootstrap interval. The rows by record count split the same 4,000 applicants, each with its own bootstrap. Tokens are the mean input tokens per call.}
\label{tab:raw}
\setlength{\tabcolsep}{4pt}
\begin{tabular*}{\linewidth}{@{\extracolsep{\fill}}lccc@{}}
\toprule
Applicants & Raw records & JevRisk & $\Delta\mathrm{AUC}$ \\
\midrule
All 4,000 (median 56 records) & 0.604 & 0.628 & $-$0.024 [$-$0.035, $-$0.011] \\
0--30 records (1,057) & 0.587 & 0.592 & $-$0.005 [$-$0.027, +0.017] \\
31--100 records (1,841) & 0.626 & 0.655 & $-$0.029 [$-$0.047, $-$0.011] \\
101--553 records (1,102) & 0.607 & 0.630 & $-$0.023 [$-$0.047, +0.001] \\
\midrule
Tokens per call & 3,229 & 1,180 & \\
\bottomrule
\end{tabular*}
\end{table}

\textbf{JevRisk with an open model.} Section~\ref{sec:setup} describes JevRisk with Qwen3.5-9B. Its thinking mode is turned off, and each question is one prompt that ends with ``Answer with Yes or No.'' Each answer is the probability of ``Yes'' against ``No'' for the first token. For Home Credit, the neutral occupation value is the median over the 575 development applicants in the sample who state an occupation, which is 0.245. For Lending Club, it is the median over the 734 development loans in the sample with a job title, which is 0.074. Table~\ref{tab:llm} splits the score into the two answers. As with \jev{}, the product is above the customer answer alone for AML and Home Credit. For Lending Club, the product is below the customer answer alone.

\begin{table}[h]
\centering
\small
\caption{\textbf{JevRisk with Qwen3.5-9B.} AUC on the 4,000 test customers of each data set. $\Delta\mathrm{AUC}$ is the AUC of the score minus that of JevRisk with \jev{}, with a 95\% bootstrap interval.}
\label{tab:llm}
\setlength{\tabcolsep}{2.5pt}
\fontsize{8.5}{9.5}\selectfont
\resizebox{\linewidth}{!}{\begin{tabular}{@{}lcccccc@{}}
\toprule
 & \multicolumn{2}{c}{AML} & \multicolumn{2}{c}{Home Credit} & \multicolumn{2}{c}{Lending Club} \\
\cmidrule(lr){2-3}\cmidrule(lr){4-5}\cmidrule(l){6-7}
Score of Qwen3.5-9B & AUC & $\Delta\mathrm{AUC}$ & AUC & $\Delta\mathrm{AUC}$ & AUC & $\Delta\mathrm{AUC}$ \\
\midrule
Product of the two answers & 0.908 & $-$0.015 [$-$0.020, $-$0.010] & 0.617 & $-$0.011 [$-$0.022, +0.000] & 0.605 & $-$0.045 [$-$0.060, $-$0.031] \\
Customer answer & 0.877 & $-$0.046 [$-$0.053, $-$0.039] & 0.608 & $-$0.020 [$-$0.033, $-$0.008] & 0.618 & $-$0.033 [$-$0.046, $-$0.019] \\
Occupation answer & 0.890 & $-$0.033 [$-$0.040, $-$0.026] & 0.599 & $-$0.029 [$-$0.042, $-$0.015] & 0.551 & $-$0.100 [$-$0.121, $-$0.080] \\
\midrule
\textbf{JevRisk (Jev-1.14.0)} & 0.923 & -- & 0.628 & -- & 0.651 & -- \\
\bottomrule
\end{tabular}\unskip}
\end{table}

\section{Baselines}
\label{app:baselines}

\textbf{Isolation forest.} The detector is fitted on all customers of the development and test sets, without labels. It thus sees more customers than JevRisk, whose thresholds use only the development customers. The detector reads the activity facts of Appendix~\ref{app:flags} or~\ref{app:credit}, each as the logarithm of one plus the value. For Home Credit, it also reads the number of children, the income, the credit amount, the instalment, the goods price and the instalment over the income. For Lending Club, it reads the facts of the credit report of Appendix~\ref{app:lending}, the loan amount, the income and the revolving balance, each as the logarithm of one plus the value, and the revolving utilisation.

\textbf{Features.} The supervised models read one row per customer. For AML, the row holds 36 features. Twenty-one are the activity facts of Appendix~\ref{app:flags}, each as the logarithm of one plus the value. Per channel, these are the counts and totals, the distinct EMT counterparties, the largest wire sent and received and the wires to or from abroad. The facts also hold the cash deposits of at least \$9,000 and below \$10,000, and money in and money out. Two more are the largest cash deposit and the largest wire to or from abroad, also as logarithms, and nine mark the foreign countries of the customer's wires, one per country. The other four features are the age, the years as a customer, the gender and the occupation. For Home Credit, the row holds 34 features. Eighteen are the facts of Appendix~\ref{app:credit}, namely seventeen from the history and the instalment over income. Five are the number of children, the income, the credit amount, the instalment and the goods price. All of these except the instalment over income enter as the logarithm of one plus the value. The other features are the age, the years employed, set to $-1$ when the data record no employment, and nine categorical fields of the application, such as the occupation, the income type and the education. A missing category is a category of its own. For Lending Club, the row holds 27 features. Twenty-two are the facts of the credit report, the loan amount, the income, the revolving balance and the revolving utilisation. The other five are the years employed and four categorical fields, namely the job title, the home ownership, the income verification and the loan purpose. The job title is free text, so the 200 most frequent titles of the development loans are categories and the other titles form one category. The supervised models thus see nearly the same information as JevRisk. For AML, the models see the same facts as JevRisk and, in addition, the gender, which JevRisk does not. For both credit data sets, both see the same profile fields, and neither sees the gender.

\textbf{Models.} The GBDT is \texttt{HistGradientBoostingClassifier} of scikit-learn with default settings, which follows LightGBM and treats the categorical features natively. Logistic regression standardises the numeric features and encodes each categorical feature one-hot. The model uses the default L2 penalty with $C=1$ and the lbfgs solver, with up to 1,000 iterations instead of the default 100. TabPFN uses version 2.0.9 of the \texttt{tabpfn} package with the public v2 classifier weights and default settings, with the categorical features marked as such. TabPFN takes at most 10,000 labelled rows, so TabPFN is not fitted on the whole development set. Each label budget uses twenty random training sets with seeds 0 to 19, and all models use the same training sets.

\section{Results for All Label Budgets}
\label{app:full}

Tables~\ref{tab:main_full}, \ref{tab:credit_full} and~\ref{tab:lending_full} report all label budgets for the AML, Home Credit and Lending Club experiments. They also show the number of high-risk customers in each labelled sample and which data each method uses.

\begin{table}[H]
\centering
\small
\caption{\textbf{AML customer risk rating for all label budgets.}}
\label{tab:main_full}
\setlength{\tabcolsep}{4pt}
\fontsize{8.5}{9.5}\selectfont
\begin{tabular*}{\linewidth}{@{\extracolsep{\fill}}lrlcc@{}}
\toprule
Method & Labels (high risk) & Customer data & AUC [95\% interval] & $\Delta\mathrm{AUC}$ \\
\midrule
Red-flag count & none & 4 percentiles & 0.830 [0.818, 0.842] & $-$0.093 [$-$0.102, $-$0.084] \\
Isolation forest & none & all, no labels & 0.898 [0.887, 0.908] & $-$0.025 [$-$0.033, $-$0.017] \\
\midrule
GBDT & 100 (3) & labelled sample & 0.803 [0.795, 0.810] & $-$0.120 [$-$0.127, $-$0.114] \\
GBDT & 300 (8) & labelled sample & 0.888 [0.880, 0.896] & $-$0.035 [$-$0.041, $-$0.029] \\
GBDT & 1,000 (28) & labelled sample & 0.906 [0.899, 0.913] & $-$0.017 [$-$0.023, $-$0.011] \\
GBDT & 3,000 (84) & labelled sample & 0.928 [0.922, 0.934] & +0.005 [$-$0.001, +0.011] \\
GBDT & 10,000 (282) & labelled sample & 0.940 [0.934, 0.946] & +0.017 [+0.011, +0.024] \\
GBDT & 38,712 (1,094) & labelled sample & 0.957 [0.951, 0.963] & +0.034 [+0.028, +0.040] \\
\midrule
Logistic regression & 100 (3) & labelled sample & 0.730 [0.724, 0.737] & $-$0.193 [$-$0.200, $-$0.185] \\
Logistic regression & 300 (8) & labelled sample & 0.843 [0.835, 0.851] & $-$0.080 [$-$0.087, $-$0.073] \\
Logistic regression & 1,000 (28) & labelled sample & 0.895 [0.887, 0.903] & $-$0.028 [$-$0.035, $-$0.021] \\
Logistic regression & 3,000 (84) & labelled sample & 0.927 [0.919, 0.934] & +0.004 [$-$0.003, +0.011] \\
Logistic regression & 10,000 (282) & labelled sample & 0.942 [0.934, 0.949] & +0.019 [+0.012, +0.026] \\
Logistic regression & 38,712 (1,094) & labelled sample & 0.949 [0.942, 0.956] & +0.026 [+0.020, +0.033] \\
\midrule
TabPFN & 100 (3) & labelled sample & 0.848 [0.841, 0.856] & $-$0.075 [$-$0.081, $-$0.068] \\
TabPFN & 300 (8) & labelled sample & 0.905 [0.897, 0.913] & $-$0.018 [$-$0.024, $-$0.011] \\
TabPFN & 1,000 (28) & labelled sample & 0.923 [0.916, 0.931] & +0.000 [$-$0.006, +0.007] \\
TabPFN & 3,000 (84) & labelled sample & 0.931 [0.924, 0.939] & +0.008 [+0.002, +0.015] \\
TabPFN & 10,000 (282) & labelled sample & 0.934 [0.927, 0.942] & +0.011 [+0.005, +0.018] \\
\midrule
JevRisk (Qwen3.5-9B) & none & 4 percentiles & 0.908 [0.899, 0.917] & $-$0.015 [$-$0.020, $-$0.010] \\
\textbf{JevRisk (Jev-1.14.0)} & none & 4 percentiles & 0.923 [0.915, 0.931] & -- \\
\bottomrule
\end{tabular*}
\end{table}

\begin{table}[H]
\centering
\small
\caption{\textbf{Credit risk assessment for all label budgets.}}
\label{tab:credit_full}
\setlength{\tabcolsep}{2pt}
\fontsize{8.5}{9.5}\selectfont
\begin{tabular*}{\linewidth}{@{\extracolsep{\fill}}lrlcc@{}}
\toprule
Method & Labels (difficulties) & Applicant data & AUC [95\% interval] & $\Delta\mathrm{AUC}$ \\
\midrule
Early-warning count & none & 2 percentiles & 0.587 [0.571, 0.603] & $-$0.041 [$-$0.055, $-$0.026] \\
Isolation forest & none & all, no labels & 0.556 [0.539, 0.573] & $-$0.072 [$-$0.090, $-$0.054] \\
\midrule
GBDT & 100 (8) & labelled sample & 0.543 [0.538, 0.548] & $-$0.085 [$-$0.102, $-$0.069] \\
GBDT & 300 (24) & labelled sample & 0.564 [0.557, 0.570] & $-$0.064 [$-$0.081, $-$0.048] \\
GBDT & 1,000 (81) & labelled sample & 0.610 [0.601, 0.619] & $-$0.018 [$-$0.034, $-$0.002] \\
GBDT & 3,000 (242) & labelled sample & 0.630 [0.620, 0.640] & +0.002 [$-$0.014, +0.018] \\
GBDT & 10,000 (807) & labelled sample & 0.665 [0.653, 0.677] & +0.037 [+0.021, +0.053] \\
GBDT & 61,694 (5,041) & labelled sample & 0.720 [0.706, 0.735] & +0.092 [+0.075, +0.110] \\
\midrule
Logistic regression & 100 (8) & labelled sample & 0.559 [0.552, 0.565] & $-$0.069 [$-$0.087, $-$0.052] \\
Logistic regression & 300 (24) & labelled sample & 0.587 [0.579, 0.595] & $-$0.041 [$-$0.058, $-$0.024] \\
Logistic regression & 1,000 (81) & labelled sample & 0.640 [0.629, 0.651] & +0.012 [$-$0.004, +0.028] \\
Logistic regression & 3,000 (242) & labelled sample & 0.673 [0.660, 0.685] & +0.045 [+0.030, +0.060] \\
Logistic regression & 10,000 (807) & labelled sample & 0.698 [0.684, 0.712] & +0.070 [+0.054, +0.086] \\
Logistic regression & 61,694 (5,041) & labelled sample & 0.714 [0.698, 0.730] & +0.086 [+0.070, +0.103] \\
\midrule
TabPFN & 100 (8) & labelled sample & 0.555 [0.549, 0.561] & $-$0.074 [$-$0.089, $-$0.058] \\
TabPFN & 300 (24) & labelled sample & 0.580 [0.573, 0.588] & $-$0.048 [$-$0.064, $-$0.032] \\
TabPFN & 1,000 (81) & labelled sample & 0.628 [0.618, 0.637] & +0.000 [$-$0.017, +0.016] \\
TabPFN & 3,000 (242) & labelled sample & 0.664 [0.652, 0.677] & +0.036 [+0.020, +0.052] \\
TabPFN & 10,000 (807) & labelled sample & 0.696 [0.682, 0.710] & +0.068 [+0.051, +0.085] \\
\midrule
JevRisk (Qwen3.5-9B) & none & 2 percentiles & 0.617 [0.600, 0.635] & $-$0.011 [$-$0.022, +0.000] \\
\textbf{JevRisk (Jev-1.14.0)} & none & 2 percentiles & 0.628 [0.611, 0.645] & -- \\
\bottomrule
\end{tabular*}
\end{table}

\begin{table}[H]
\centering
\small
\caption{\textbf{Credit risk assessment on Lending Club for all label budgets.} The test loans were issued in 2015, after the development loans.}
\label{tab:lending_full}
\setlength{\tabcolsep}{3pt}
\fontsize{8.5}{9.5}\selectfont
\begin{tabular*}{\linewidth}{@{\extracolsep{\fill}}lrlcc@{}}
\toprule
Method & Labels (bad) & Applicant data & AUC [95\% interval] & $\Delta\mathrm{AUC}$ \\
\midrule
Early-warning count & none & 4 percentiles & 0.574 [0.557, 0.591] & $-$0.077 [$-$0.094, $-$0.059] \\
Isolation forest & none & all, no labels & 0.544 [0.526, 0.562] & $-$0.107 [$-$0.129, $-$0.085] \\
\midrule
GBDT & 100 (13) & labelled sample & 0.552 [0.545, 0.559] & $-$0.099 [$-$0.113, $-$0.084] \\
GBDT & 300 (40) & labelled sample & 0.570 [0.563, 0.579] & $-$0.080 [$-$0.095, $-$0.066] \\
GBDT & 1,000 (132) & labelled sample & 0.591 [0.582, 0.600] & $-$0.060 [$-$0.074, $-$0.046] \\
GBDT & 3,000 (396) & labelled sample & 0.609 [0.599, 0.620] & $-$0.042 [$-$0.055, $-$0.028] \\
GBDT & 10,000 (1,319) & labelled sample & 0.636 [0.623, 0.648] & $-$0.015 [$-$0.028, $-$0.002] \\
GBDT & 262,992 (34,693) & labelled sample & 0.678 [0.662, 0.693] & +0.027 [+0.012, +0.041] \\
\midrule
Logistic regression & 100 (13) & labelled sample & 0.571 [0.563, 0.578] & $-$0.080 [$-$0.094, $-$0.066] \\
Logistic regression & 300 (40) & labelled sample & 0.594 [0.585, 0.604] & $-$0.056 [$-$0.070, $-$0.042] \\
Logistic regression & 1,000 (132) & labelled sample & 0.629 [0.616, 0.641] & $-$0.022 [$-$0.035, $-$0.009] \\
Logistic regression & 3,000 (396) & labelled sample & 0.653 [0.638, 0.667] & +0.002 [$-$0.011, +0.015] \\
Logistic regression & 10,000 (1,319) & labelled sample & 0.664 [0.649, 0.679] & +0.014 [+0.000, +0.028] \\
Logistic regression & 262,992 (34,693) & labelled sample & 0.681 [0.664, 0.698] & +0.031 [+0.016, +0.045] \\
\midrule
TabPFN & 100 (13) & labelled sample & 0.571 [0.563, 0.579] & $-$0.080 [$-$0.094, $-$0.066] \\
TabPFN & 300 (40) & labelled sample & 0.597 [0.586, 0.607] & $-$0.054 [$-$0.068, $-$0.040] \\
TabPFN & 1,000 (132) & labelled sample & 0.630 [0.617, 0.643] & $-$0.021 [$-$0.033, $-$0.008] \\
TabPFN & 3,000 (396) & labelled sample & 0.656 [0.641, 0.672] & +0.006 [$-$0.007, +0.019] \\
TabPFN & 10,000 (1,319) & labelled sample & 0.667 [0.651, 0.682] & +0.016 [+0.003, +0.029] \\
\midrule
JevRisk (Qwen3.5-9B) & none & all, no labels & 0.605 [0.588, 0.623] & $-$0.045 [$-$0.060, $-$0.031] \\
\textbf{JevRisk (Jev-1.14.0)} & none & all, no labels & 0.651 [0.634, 0.667] & -- \\
\bottomrule
\end{tabular*}
\end{table}

\section{Precision and Recall at the Top of the Ranking}
\label{app:pr}

A bank reviews only the top of the ranking, so Table~\ref{tab:pr} adds two metrics for the top. Average precision (AP) is the area under the precision-recall curve. Recall@$k$\% is the percentage of all high-risk customers that rank in the top $k$\%. A random ranking has an AP equal to the percentage of high-risk customers and a recall@$k$\% of $k$\%. Unlike the AUC, both metrics depend on the percentage of high-risk customers. The test sample holds as many high-risk customers as other customers. Each sampled other customer therefore gets the weight $N_{\mathrm{neg}}/N_{\mathrm{pos}}$ of the test set, about 35 for AML, 11 for Home Credit and 5.7 for Lending Club, so that the weighted sample stands for the whole test set. Many customers share an indicator count. In the AUC, a high-risk customer and a low-risk customer with tied scores count half. In AP and recall, a group of tied scores at the cut counts in proportion to its part above the cut, which is the expected value when ties are broken at random. All metrics use the scores of the main results, so the table takes no further calls. The table covers all methods and label budgets of Tables~\ref{tab:main_full}, \ref{tab:credit_full} and~\ref{tab:lending_full}, with the same training sets for the supervised models. Each low-risk customer in the sample stands for about 35, 11 or 5.7 customers of the test set. One such customer near the top of the ranking thus counts as 35, 11 or 5.7 false alarms. The metrics therefore change a lot across the bootstrap samples, and the intervals are wide.

\begin{table}[H]
\centering
\small
\caption{\textbf{Precision and recall at the top of the ranking on the test samples.} AP is the average precision, and Recall@$k$\% is the percentage of all high-risk customers that rank in the top $k$\% of the test set. For credit, the high-risk customers are the applicants with payment difficulties for Home Credit and the bad loans for Lending Club. Other customers are weighted to the percentage of high-risk customers in the test set. For the supervised models, each value is the mean over the twenty training sets, and the rows labelled all use the whole development set. $^{\ast}$ and $^{\dagger}$ mark a method below or above JevRisk with \jev{}, with a paired 95\% bootstrap interval of the difference that excludes zero.}
\label{tab:pr}
\setlength{\tabcolsep}{2.5pt}
\fontsize{8.5}{9.5}\selectfont
\resizebox{\linewidth}{!}{\begin{tabular}{@{}lrccccccccc@{}}
\toprule
 & & \multicolumn{3}{c}{AML} & \multicolumn{3}{c}{Home Credit} & \multicolumn{3}{c}{Lending Club} \\
\cmidrule(lr){3-5}\cmidrule(lr){6-8}\cmidrule(l){9-11}
Method & Labels & AP & \begin{tabular}{@{}c@{}}Recall\\@5\%\end{tabular} & \begin{tabular}{@{}c@{}}Recall\\@10\%\end{tabular} & AP & \begin{tabular}{@{}c@{}}Recall\\@5\%\end{tabular} & \begin{tabular}{@{}c@{}}Recall\\@10\%\end{tabular} & AP & \begin{tabular}{@{}c@{}}Recall\\@5\%\end{tabular} & \begin{tabular}{@{}c@{}}Recall\\@10\%\end{tabular} \\
\midrule
Random ranking & none & 0.028 & 0.050 & 0.100 & 0.080 & 0.050 & 0.100 & 0.149 & 0.050 & 0.100 \\
Indicator count & none & 0.169$^{\ast}$ & 0.377$^{\ast}$ & 0.580$^{\ast}$ & 0.103$^{\ast}$ & 0.081$^{\ast}$ & 0.154$^{\ast}$ & 0.180$^{\ast}$ & 0.078$^{\ast}$ & 0.145$^{\ast}$ \\
Isolation forest & none & 0.264 & 0.464 & 0.696 & 0.099$^{\ast}$ & 0.084$^{\ast}$ & 0.134$^{\ast}$ & 0.165$^{\ast}$ & 0.053$^{\ast}$ & 0.117$^{\ast}$ \\
\midrule
GBDT & 100 & 0.124$^{\ast}$ & 0.241$^{\ast}$ & 0.425$^{\ast}$ & 0.094$^{\ast}$ & 0.066$^{\ast}$ & 0.125$^{\ast}$ & 0.176$^{\ast}$ & 0.067$^{\ast}$ & 0.129$^{\ast}$ \\
GBDT & 300 & 0.298 & 0.512 & 0.704 & 0.102$^{\ast}$ & 0.076$^{\ast}$ & 0.143$^{\ast}$ & 0.185$^{\ast}$ & 0.075$^{\ast}$ & 0.141$^{\ast}$ \\
GBDT & 1,000 & 0.333 & 0.514 & 0.706 & 0.123 & 0.104 & 0.181 & 0.194$^{\ast}$ & 0.078$^{\ast}$ & 0.149$^{\ast}$ \\
GBDT & 3,000 & 0.459$^{\dagger}$ & 0.602$^{\dagger}$ & 0.769$^{\dagger}$ & 0.134 & 0.111 & 0.201 & 0.205$^{\ast}$ & 0.083$^{\ast}$ & 0.160$^{\ast}$ \\
GBDT & 10,000 & 0.562$^{\dagger}$ & 0.682$^{\dagger}$ & 0.810$^{\dagger}$ & 0.162$^{\dagger}$ & 0.142$^{\dagger}$ & 0.243$^{\dagger}$ & 0.223 & 0.091 & 0.174 \\
GBDT & all & 0.606$^{\dagger}$ & 0.762$^{\dagger}$ & 0.879$^{\dagger}$ & 0.201$^{\dagger}$ & 0.173$^{\dagger}$ & 0.293$^{\dagger}$ & 0.259$^{\dagger}$ & 0.110 & 0.213 \\
\midrule
Logistic regression & 100 & 0.226$^{\ast}$ & 0.391$^{\ast}$ & 0.519$^{\ast}$ & 0.102$^{\ast}$ & 0.078$^{\ast}$ & 0.144$^{\ast}$ & 0.186$^{\ast}$ & 0.074$^{\ast}$ & 0.142$^{\ast}$ \\
Logistic regression & 300 & 0.279 & 0.472 & 0.632$^{\ast}$ & 0.112$^{\ast}$ & 0.090 & 0.162$^{\ast}$ & 0.198$^{\ast}$ & 0.081$^{\ast}$ & 0.155$^{\ast}$ \\
Logistic regression & 1,000 & 0.364 & 0.564$^{\dagger}$ & 0.726 & 0.138 & 0.119 & 0.209 & 0.217$^{\ast}$ & 0.090 & 0.169 \\
Logistic regression & 3,000 & 0.512$^{\dagger}$ & 0.662$^{\dagger}$ & 0.812$^{\dagger}$ & 0.157$^{\dagger}$ & 0.136$^{\dagger}$ & 0.236$^{\dagger}$ & 0.236 & 0.097 & 0.181 \\
Logistic regression & 10,000 & 0.585$^{\dagger}$ & 0.718$^{\dagger}$ & 0.848$^{\dagger}$ & 0.179$^{\dagger}$ & 0.154$^{\dagger}$ & 0.262$^{\dagger}$ & 0.245 & 0.101 & 0.185 \\
Logistic regression & all & 0.614$^{\dagger}$ & 0.736$^{\dagger}$ & 0.868$^{\dagger}$ & 0.195$^{\dagger}$ & 0.164$^{\dagger}$ & 0.276$^{\dagger}$ & 0.258$^{\dagger}$ & 0.099 & 0.194 \\
\midrule
TabPFN & 100 & 0.246$^{\ast}$ & 0.423$^{\ast}$ & 0.601$^{\ast}$ & 0.098$^{\ast}$ & 0.069$^{\ast}$ & 0.132$^{\ast}$ & 0.184$^{\ast}$ & 0.072$^{\ast}$ & 0.135$^{\ast}$ \\
TabPFN & 300 & 0.336 & 0.541$^{\dagger}$ & 0.733 & 0.109$^{\ast}$ & 0.083$^{\ast}$ & 0.153$^{\ast}$ & 0.196$^{\ast}$ & 0.078$^{\ast}$ & 0.146$^{\ast}$ \\
TabPFN & 1,000 & 0.435$^{\dagger}$ & 0.636$^{\dagger}$ & 0.798$^{\dagger}$ & 0.133 & 0.112 & 0.196 & 0.218$^{\ast}$ & 0.088 & 0.170 \\
TabPFN & 3,000 & 0.521$^{\dagger}$ & 0.702$^{\dagger}$ & 0.821$^{\dagger}$ & 0.157$^{\dagger}$ & 0.135$^{\dagger}$ & 0.232$^{\dagger}$ & 0.237 & 0.100 & 0.188 \\
TabPFN & 10,000 & 0.546$^{\dagger}$ & 0.732$^{\dagger}$ & 0.829$^{\dagger}$ & 0.182$^{\dagger}$ & 0.156$^{\dagger}$ & 0.272$^{\dagger}$ & 0.247 & 0.103 & 0.195 \\
\midrule
JevRisk (Qwen3.5-9B) & none & 0.291 & 0.463 & 0.696 & 0.133 & 0.110 & 0.204 & 0.198$^{\ast}$ & 0.073$^{\ast}$ & 0.154$^{\ast}$ \\
\textbf{JevRisk (Jev-1.14.0)} & none & 0.308 & 0.493 & 0.715 & 0.129 & 0.106 & 0.196 & 0.237 & 0.104 & 0.187 \\
\quad 95\% interval &  & {\scriptsize [0.265, 0.374]} & {\scriptsize [0.444, 0.541]} & {\scriptsize [0.670, 0.753]} & {\scriptsize [0.119, 0.146]} & {\scriptsize [0.088, 0.123]} & {\scriptsize [0.169, 0.216]} & {\scriptsize [0.220, 0.258]} & {\scriptsize [0.087, 0.121]} & {\scriptsize [0.166, 0.208]} \\
\bottomrule
\end{tabular}\unskip}
\end{table}

\section{Compute and Cost}
\label{app:cost}

For AML, the main result takes 4,000 \jev{} calls on the test sample and reads 4.5 million tokens. Table~\ref{tab:stability} takes 1,044 further calls for the customers whose flags changed. For Home Credit, the main result takes 4,000 calls on the test sample, which read 4.7 million tokens, and 800 calls on development applicants for the neutral value. For Lending Club, the main result takes 4,000 calls on the test sample, which read 5.5 million tokens, and 800 calls on development loans for the neutral value. The raw-record ablation of Appendix~\ref{app:ablation} takes 4,800 calls and 15.5 million tokens, and the ablation without indicators takes 13,309 calls and 9.5 million tokens. JevRisk with Qwen3.5-9B takes 25,309 prompts and runs in about four hours on three RTX 2080 Ti GPUs. On the same GPUs, processing 13,309 requests takes about two hours with Clef-Flash and four hours with Jev-Omni. The API charges only for input tokens, at \$0.042 per million as TypeSafe listed in September 2026 \citep{typesafe2026jev}. At about 1,120, 1,180 and 1,370 tokens per call, rating all 195,789 AML customers once costs about \$9, rating all 307,511 Home Credit applicants about \$15, and rating all 546,018 Lending Club loans about \$31. The GBDT, logistic regression and the isolation forest run on a CPU in minutes. TabPFN runs on an RTX 2080 Ti GPU. All TabPFN and logistic regression fits take about four minutes per data set.

\section{Limitations}
\label{app:limits}

\textbf{Data.} For AML, the data and labels are synthetic. Moreover, the labels are a customer risk rating by the bank, not confirmed laundering cases or filed suspicious transaction reports. The transactions carry no dates, so we cannot apply time-based scenarios, such as FINTRAC's 24-hour rule, which adds up the cash deposits made within 24 hours, or velocity rules, which flag many transactions in a short time. For Home Credit, the data are real, but the label marks payment difficulties on one loan, not a default under a regulatory definition. For Lending Club, the data hold only approved loans, so the applicants who were declined are missing. Only Lending Club has dates, so only its test set comes after its development set. For credit, the AUC of all methods is modest, and we leave out the external credit scores and the FICO score that a lender would use.

\textbf{Method and evaluation.} Peer-group thresholds use the unlabelled activity of the development customers, which any bank records. JevRisk is therefore label-free but not free of other customers' data. The questions to \jev{} need expert design. An expert chooses the risk factors that the questions ask about, such as the occupation or the recent credit seeking, and the wording of a question can change what \jev{} weighs. We study \jev{} in one API version, and its outputs can change across versions. We do not test JevRisk for bias across protected groups, such as groups by gender, age or ethnicity. The occupation question in particular may act as a proxy for such groups, so a bank would need fairness tests before use. Finally, we score a sample of 4,000 test customers per data set, not the whole test set. We call two methods on par when the 95\% interval of their difference contains zero. This means that the sample shows no clear difference, not that the two methods are equivalent. The intervals reflect only the sampling of the test customers, not that of the twenty training sets.

\end{document}